\documentclass{article}

\PassOptionsToPackage{numbers, compress}{natbib}

\usepackage[preprint]{neurips_2023}

\usepackage[utf8]{inputenc} 
\usepackage[T1]{fontenc}    
\usepackage{hyperref}       
\usepackage{url}            
\usepackage{booktabs}       
\usepackage{amsfonts}       
\usepackage{nicefrac}       
\usepackage{microtype}      
\usepackage{xcolor, colortbl}         

\usepackage{tabularx}
\usepackage{multirow}  
\usepackage{bm}
\usepackage{siunitx}
\usepackage{wrapfig}
\usepackage{float}
\usepackage{graphicx}
\usepackage{subfigure}

\DeclareSIUnit\angstrom{\text {Å}}
\definecolor{gray_others}{gray}{0.9}
\definecolor{gray_shape}{gray}{0.83}

\title{Reinforcement Learning-Driven Linker Design via Fast Attention-based Point Cloud Alignment}

\author{%
  Rebecca M.~Neeser
  \\
  VantAI\\
  New York, NY 10036 \\
  \texttt{rebecca@vant.ai} \\
  \And
  Mehmet Akdel \\
  VantAI\\
  New York, NY 10036 \\
  \texttt{mehmet@vant.ai} \\
  \AND
  Daniel Kovtun \\
  VantAI\\
  New York, NY 10036 \\
  \texttt{danny@vant.ai} \\
  \And
  Luca Naef \\
  VantAI\\
  New York, NY 10036 \\
  \texttt{luca@vant.ai} \\
}

\begin{document}

\maketitle

\begin{abstract}
Proteolysis-Targeting Chimeras (PROTACs) represent a novel class of small molecules which are designed to act as a bridge between an E3 ligase and a disease-relevant protein, thereby promoting its subsequent degradation. PROTACs are composed of two protein binding "active" domains, linked by a "linker" domain. The design of the linker domain is challenging due to geometric and chemical constraints given by its interactions, and the need to maximize drug-likeness. To tackle these challenges, we introduce ShapeLinker, a method for \textit{de novo} design of linkers. It performs fragment-linking using reinforcement learning on an autoregressive SMILES generator. The method optimizes for a composite score combining relevant physicochemical properties and a novel, attention-based point cloud alignment score. This new method successfully generates linkers that satisfy both relevant 2D and 3D requirements, and achieves state-of-the-art results in producing novel linkers assuming a target linker conformation. This allows for more rational and efficient PROTAC design and optimization. Code and data are available at \url{https://github.com/aivant/ShapeLinker}.
\end{abstract}

\section{Introduction}
\label{sec:intro}
Most small-molecule drugs act by interfering with a disease-causing protein of interest~(POI) through inhibition or activation of its function via a functional binding site. However, approximately 80\% of the human proteome lacks such a binding site, requiring alternative drug modalities.\cite{crews_targeting_2010} Proteolysis-targeting chimeras~(PROTACs) can act on these "undruggable" targets.\cite{sakamoto2001protacs} PROTACs exhibit their mode of action by binding two proteins -- an enzyme of the class of E3 ligases and the POI. This induced proximity enables the ubiquitination of the POI by the E3 ligase, which marks the POI for degradation.\cite{lai_induced_2017} Additionally, this catalytic mode of action enables a given PROTAC molecule to degrade multiple molecules of a given POI, allowing for sub-stochiometric concentrations to achieve therapeutic effects. PROTACs are hetero-bifunctional small molecules consisting of an anchor fragment binding the E3 ligase, a warhead targeting the POI, and a linker joining these two ligands. The combination of these fragments results in a small molecule of relatively large size (700-1100~Da) compared to traditional small molecule drugs (\(<500\)~Da), which poses additional challenges related to e.g. lipophilicity or metabolic stability.\cite{an_small-molecule_2018} During PROTAC discovery campaigns, the linker is a key lever for optimization and frequently iterated to optimize both chemical properties such as hydrophobicity, solubility, and overall degradation efficiency.\cite{bemis_unraveling_2021} \par 

The inherent complexity of the ternary complex, where the linker does not occupy a traditional pocket, makes rational design of PROTAC linkers particularly challenging. Machine learning~(ML)-based linker generation methods enable rational design of novel linkers with a significantly lower computational cost than traditional physics-based simulations. Existing generative models for fragment-linking have limited practical utility as they have either been based on only 2D representations, or do not allow for explicit, modular optimization towards desired linker chemical spaces (e.g., rigidity, physicochemical properties, limiting branching). \cite{bemis_unraveling_2021}. However, when designing PROTACs, taking both into consideration simultaneously is required. While ternary complex formation between the E3 ligase and POI components does not guarantee a functional outcome, accumulating evidence suggests that the efficiency, stability, and spatial arrangement of ternary complex distributions induced by a given molecule are critical to driving degradation.\cite{5T35_xtal, law_discovery_2021, 6BN7_6BOY_xtal} Since there is less room to influence the ternary complex via modification of the individual cognate ligands, designing linkers that can effectively stabilize desired ternary complex conformations is crucial.\cite{6BN7_6BOY_xtal, chamberlain_development_2019, lv_development_2021, bemis_unraveling_2021} \par 

This work aims to address these challenges by introducing a novel 3D shape-conditioned linker generation method, ShapeLinker, which allows multi-parameter-optimization using reinforcement learning~(RL) to steer the design efforts in the desired chemical space. We combine advantages of previous 2D methods (modular optimization) and introduce a novel, fast attention-based point cloud alignment method for conditioning the generation on geometric features. This new shape alignment method allows us to optimize to a reference linker shape known to stabilize a productive ternary complex. Our efforts mainly contribute to the linker design for the drug modality of PROTACs and their specific requirements. This method thus enables efficient lead optimization against predicted or known structures of E3-POI interfaces or known binders in other ternary configurations.

\section{Related Work}
\label{sec:related_work}
\textit{De novo} linker design through generative models has primarily been addressed in the context of fragment-based drug design~(FBDD).\cite{sheng2013fragment} However, such methods may not be suited to the linker design for large structures such as PROTACs, as they aim at connecting substantially smaller fragments. Both FBDD and \textit{de novo} linker design will be discussed in the following sections.\par 

Various fragment-linking methods generate molecules in 2D. SyntaLinker~\cite{yang_syntalinker_2020} is a transformer-based FBDD method viewing the fragment-linking as a "sentence-completion"~\cite{zweig2012computational} task using the string-based SMILES (Simplified molecular-input line-entry system)~\cite{weininger1988smiles} representation that can be conditioned on physicochemical properties. \citet{feng_syntalinker-hybrid_2022} introduced SyntaLinker-Hybrid improving target-specificity through transfer learning and PROTAC-RL~\cite{zheng_accelerated_2022} adapts SyntaLinker to specifically design linkers for PROTACs optimizing for linker length, logP and a custom pharmacokinetic~(PK) score. Link-INVENT~\cite{guo_link-invent_2022} is a recurrent neural network~(RNN) based SMILES generator building on work by \citet{fialkova_libinvent_2022} and \citet{blaschke_reinvent_2020}. The published prior for Link-INVENT was trained on fragmented drug-like molecules from ChEMBL. We base our work in this paper on Link-INVENT due to its ability to perform multi-parameter optimization through RL, allowing the user to steer the generation towards the desired chemical space. Furthermore, the use of the aforementioned prior enables generation of syntactically valid SMILES and autoregressive models are known to exhibit low inference time. While all previous methods use SMILES as molecular representation, GraphINVENT~\cite{nori_novo_nodate} makes use of a graph-based representation for the \textit{de novo} design of full PROTACs and optimizes through RL using a degradation predictor. However, since this approach attempts to design not only the linker but the whole PROTAC this model is more suitable to the hit finding stage where anchor and warhead are unknown.\par 
None of the aforementioned methods take geometry into account, which is thought to contribute substantially to the potency and thus efficacy of a drug.\cite{ramirez_computational_2016, chamberlain_development_2019} This was first addressed by \citet{delinker} proposing the graph-based DeLinker, which inputs limited geometric information as constraints. DEVELOP~\cite{delinker_develop} extends the method to include pharmacophore information and \citet{fleck_decoupled_2022} attempted at improving the robustness of the predicted coordinates. \citet{huang_3dlinker_2022} proposed 3DLinker, which utilizes more explicit geometry information and is based on an equivariant graph variational autoencoder. In our experience both DeLinker and 3DLinker often do not produce chemically sensible linkers, especially for longer linker fragments. Presumably, this is due to the fact that the chemical space the models were trained upon covers traditionally drug-like molecules and does not include the long distances required by the exotic nature of some PROTAC linkers. \citet{squid} introduced SQUID for FBDD, which leverages shape-conditioning to link fragments by generating molecules similar to a query in shape but diverse in their 2D chemistry. However, this method is not suitable to linker generation as it only connects each fragment by one rotatable bond. Joining the recent surge in diffusion models, \citet{difflinker} proposed DiffLinker, which predicts atom types and coordinates of linkers using atomic point clouds. DiffLinker enables protein pocket-conditioning and achieves state-of-the art performance on 3D metrics, albeit with a relatively high inference time. While LINK-invent and other RL-based PROTAC design methods so far did not consider 3D geometry, REINVENT for small molecule design was shown to allow for geometry conditioning using ROCS (Rapid Overlay of Chemical Structures)~\cite{rocs}, suggesting 3D conditioning may also work for linker design.\cite{papadopoulos_novo_2021}. ROCS assesses 3D shape and pharmacophore similarity simultaneously. However, ROCS requires an OpenEye license and is not fast enough to scale to our RL needs, which is also the case for the widely used RANSAC method. We developed a novel approach to perform alignment on dense surface point clouds with a multi-head attention architecture. The shape alignment allows us to guide the generation towards linker shapes known to stabilize productive ternary complex poses. This scalable aligner takes advantage of GPUs, takes 230~ms per small molecule pair, and improves over global RANSAC alignment. In order to perform the alignment, conformers need to be generated for all the sampled SMILES, which is the time bottleneck during training of ShapeLinker.
\par 

\begin{figure}[t]
    \centering
    \includegraphics[width=0.8\textwidth]{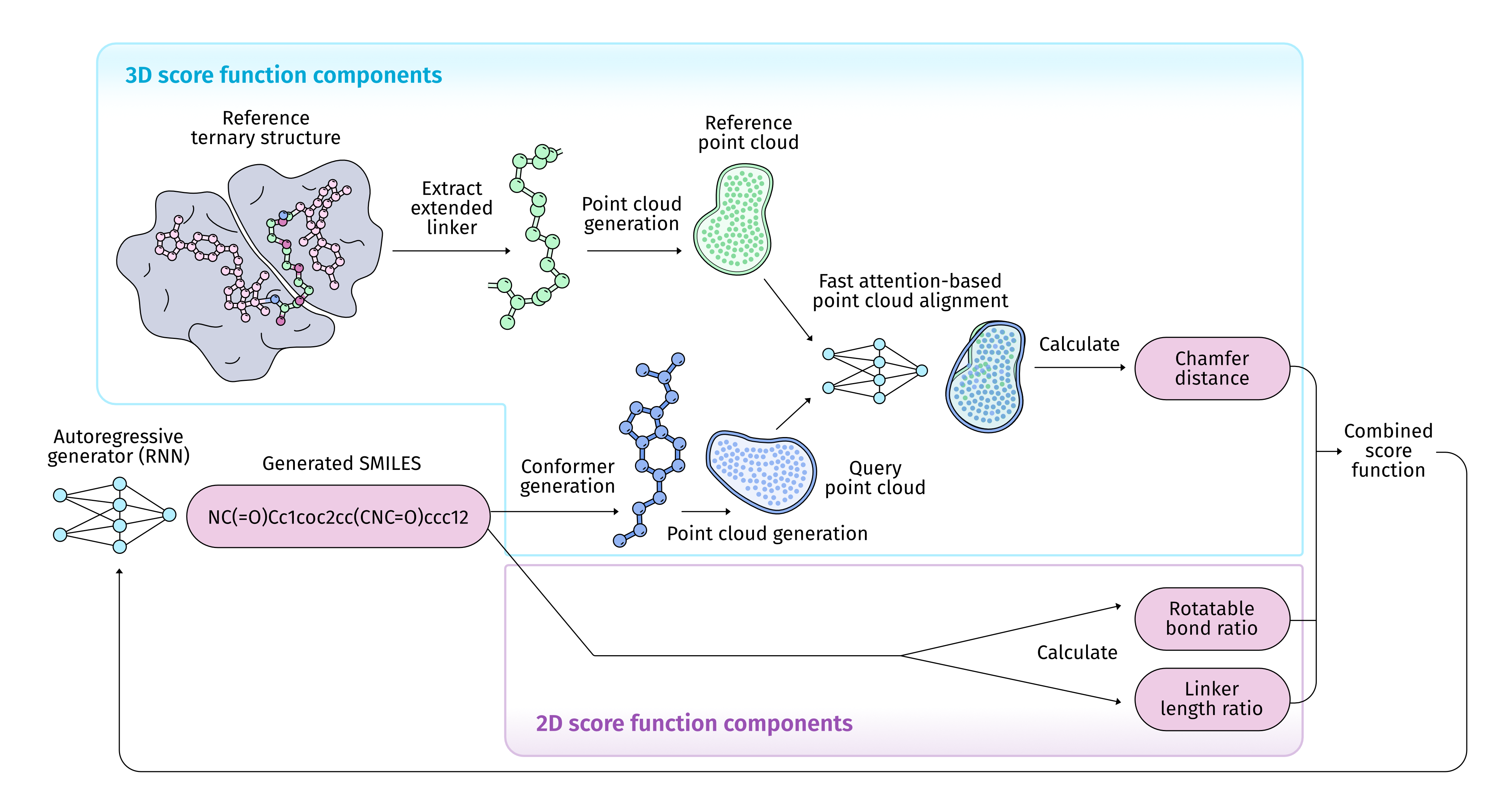}
    \caption{Schematic overview of ShapeLinker. The surface point clouds of generated molecules are aligned and scored using a trained multi-head attention alignment model.}
    \label{fig:RL-scheme}
\end{figure}

\section{Methods}
\label{sec:methods}

\subsection{Shape alignment}
\label{sec:method_alignment}

Point clouds have been successfully used as input to attention-based and transformer-based neural network architecture for applications in molecule generation.\cite{zhao_point_2021, qi_pointnet_2017} Taking advantage of the introduction of a differentiable Kabsch alignment~\cite{ganea_independent_2022,stark_equibind_2022}, we build on these ideas to perform global point cloud alignment. This is applied to point clouds derived from molecular surfaces, as these are more dense and more relevant to the task at hand, i.e. interfacial binding.\par 

\begin{figure}[b]
    \centering
    \includegraphics[width=0.8\textwidth]{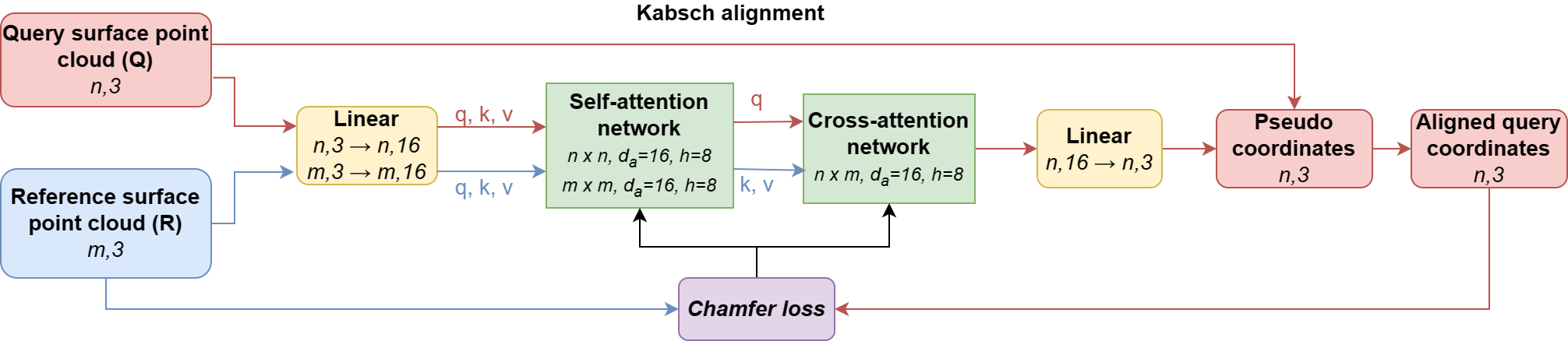}
    \caption{Multi-head attention model for global point cloud alignment.}
    \label{fig:model-scheme}
\end{figure}

Our shape alignment approach takes pairs of surface point clouds of molecules as input, one query and one target. Surface point clouds are generated using the KeOps library~\cite{feydy_fast_2020}, which was used by previous molecular modeling tasks such as dMaSIF~\cite{sverrisson_fast_2021}. This process is described in more detail in Appendix~\ref{sec:SI_pcgen}. These point cloud coordinates are translated such that their centroids are at the origin. We then train a deep learning model composed of a multi-head self-attention layer which acts individually on each point cloud, and a multi-head cross-attention layer which acts on the query-target pair. The attention layers apply full attention between all points in the input point clouds, leading to a global alignment with context from the entire molecules. The model predicts query pseudo-coordinates to which the query molecule is superposed, using a differentiable Kabsch algorithm~\cite{ganea_independent_2022,stark_equibind_2022}, to obtain the aligned pose (see \ref{sec:SI_shape_arch} for architecture details). We use a normalized L2 version of Chamfer loss~\cite{fan_point_2017}, which is a measure of global distance between the aligned query coordinates and target coordinates. We define the Chamfer distance~(CD) between two point sets \(A\) and \(B\) as:
\begin{equation}
  \text{CD}(A, B) = \frac{\sum_i \min_j(\|a_i - b_j\|^2_2) + \sum_j \min_i(\|a_i - b_j\|^2_2)}{\left|A\right| + \left|B\right|}
\end{equation}

The model was trained for 50~epochs, achieving an improvement over the RANSAC distance of over one on the validation set.\cite{ransac_open3d} The model resulting from this approach was used in all further shape alignment tasks including for scoring during RL (\textit{vide infra}).

\subsection{ShapeLinker: Geometry-conditioned linker design}
\label{sec:method_linkinvent_shape}

ShapeLinker, our geometry-conditioned method for generating SMILES linking two input fragments, is based on Link-INVENT by \citet{guo_link-invent_2022}. Link-INVENT is made up of an encoder-decoder architecture each consisting of RNNs with hidden size~256 connected by three~long short-term memory cells (LSTM).\cite{guo_link-invent_2022} We follow Link-INVENT's implementation to iteratively update the agent network through policy-based RL in the following fashion:
\begin{enumerate}
    \item Sample batch size of linkers based on the current agent.
    \item Score the generated molecules using a custom scoring function.
    \item Compute the loss and update the agent's policy.
\end{enumerate}

The loss~\(J(\theta)\) is defined as the difference between augmented and posterior likelihoods.\cite{fialkova_libinvent_2022} The augmented log likelihood is defined as follows, with \(\pi\) the probability of sampling a token based on the already present token sequence, \(S(x)\) the scoring function and a scalar factor~\(\sigma\) of 120:
\begin{equation}
    \log\pi_{\text{augmented}}=\log\pi_{\text{prior}}+\sigma S(x)
\end{equation}

The augmented log likelihood is then subtracted by the log likelihood of the current agent as follows:(

\begin{equation}
    J(\theta)=(\log\pi_{\text{augmented}}-\log\pi_{\text{agent}})^2
\end{equation}

The scoring function~\(S(x)\) to adjust the prior log-likelihood defines the key objectives for the parameter optimization and the various scores are combined in a weighted mean as follows:

\begin{equation}
    S(x)= \left[\prod_i C_i(x)^{w_i}\right]^{1/\sum_iw_i}
    \label{eq:scoring_fct}
\end{equation}
with \(C_i\) the individual score and \(w_i\) the weight of the \textit{i}th component. The composite scoring function used in ShapeLinker consists of three scores: 
\begin{enumerate}
    \item \textbf{Shape alignment} (\(w_1=3\)): Chamfer distance~(CD) between sample~\(x\) and the reference crystal structure pose determined by the shape alignment model. The alignment is carried out on the level of the extended linker (\textit{vide supra}) with 16~conformers generated for each linker and the smallest distance of those corresponds to the raw score for sample~\(x\). The raw CD is subsequently scaled using a reverse sigmoid transformation with a upper bound of 3.5 (low score) a lower bound of 0 (high score) and a steepness of 0.25.
    \item \textbf{Ratio of rotatable bonds} (\(w_2=1\)): number of rotatable bonds divided by the total number of bonds in the linker. This score corresponds to rigidity of the linker and and a score of 1 is awarded if the sample \(x\) achieved a value in \([0, 30]\) (high rigidity) and 0 in any other case.
    \item \textbf{Linker length ratio} (\(w_3=1\)): number of bonds between attachment atoms divided by the number of bonds in the longest graph path. This score controls for branching and a score of 1 is awarded if the sample \(x\) had a ratio of 100 (no branching) and 0 in any other case.
\end{enumerate}

Scores 2 and 3 were already implemented in Link-INVENT while score 1 is a new contribution of this paper. The shape alignment contribution is weighted the highest as learning such a complex property is substantially more challenging than the other two and was also thought to be more important for this task. Generated molecules are only scored for training and not during inference. Lastly, the scoring is also affected by a diversity filter as implemented in REINVENT~\cite{blaschke_reinvent_2020}, which allows penalization of recurring Murcko scaffolds in order to explore a new chemical space. This parameter optimization was carried out separately for every investigated system. 5,000~molecules for subsequent evaluation were sampled from the last agent for every system, applying a temperature scaling (\(T=1.5\)) of the logits to lower the model's confidence and in turn increase uniqueness.

\subsection{Data}
\label{sec:methods_data}
Two datasets were used: PROTAC-DB~\cite{protacdb}, which contains a large collection of publicly available data on PROTACs including both crystallized and modelled ternary complexes, and a hand-selected set of ten well known crystal structures of ternary complexes extracted from the Protein Data Bank (PDB)~\cite{PDB}. The data processing is detailed in Appendix~\ref{sec:SI_data}. \par 

PROTAC-DB is used for both training the shape alignment method, by taking a random selection, and in its entirety (3,182 after filtering) as a reference for assessing the novelty metrics. In order to reduce the computation cost, the shape alignment is done using only the respective linker with small fragments extending into both ligand fragments, rather than the full PROTAC structure. The extension of the linker to the individual fragments is critical, as the optimal geometry of the linker will be dictated by the degrees of freedom of the fully-constructed PROTAC molecule, rather than the linker in isolation. The ten ternary complexes (PDB IDs: 5T35, 7ZNT, 6BN7, 6BOY, 6HAY, 6HAX, 7S4E, 7JTP, 7Q2J, 7JTO) all have binding PROTACs that were optimized in individual structure-based drug studies and cover a diverse range of shapes with the shortest path between anchor and warhead ligands ranging from 3~atoms (7JTP) to 13~atoms (7JTO and 6BN7) while 3D distances between the anchoring atoms range from 4.73~\si{\angstrom} (7JTP) to 12.86~\si{\angstrom} (6BOY). We include these in the training of the shape alignment model as queries. Subsequently, we train an RL agent for each structure as a benchmark of (conditional) linker design methods. 

\subsection{Evaluation}
\label{sec:methods_eval}
\subsubsection{Constrained embedding}
\label{sec:const_embed}
In order to establish a fair comparison at the 3D level, we applied a constrained embedding algorithm to the unique SMILES strings generated by all three methods - ShapeLinker, DiffLinker and Link-INVENT. Only molecules passing the 2D filters (cf. section \ref{sec:methods_metrics}), a synthetic accessibility~(SA) score~\cite{SA_score} for the linker fragment of less than 4 and those with no formal charges were taken into consideration. The constrained embedding process attempts to create conformers of the PROTAC molecule given fixed atom coordinates for the non-linker substructures as constraints. which are extracted directly from the crystal structure. Using coordinate constraints to generate 3D conformations can lead to highly strained conformations since the rotatable bonds of the substructures are held fixed. To refine the conformations, we carry out several optimizations to minimize the strain energy. The constrained embedding pipeline, including post-processing, is as follows:
\begin{enumerate}
    \item Constrained embedding with crystal structures of anchor and warhead as constraints with subsequent energy minimization of the linker using RDKit~\cite{rdkit}
    \item Geometry optimization and energy minimization of the whole molecule with the MMFF94s force field using a steepest-descent algorithm implemented in OpenBabel~\cite{openbabel}.
    \item Empirical scoring minimization of the small molecule in the context of the rigid protein, using smina~\cite{smina}.
    \item Selection of the best conformer per molecule based on the combination of normalized (min-max scaling) vinardo score and RMSD.
\end{enumerate}

\subsubsection{Metrics}
\label{sec:methods_metrics}
An array of various evaluation metrics are reported. First, measures assessing the generative properties of the methods are calculated according to GuacaMol.\cite{brown_guacamol_2019} These include validity, uniqueness and novelty (with PROTAC-DB as reference), where the latter two do not take stereochemistry into consideration. Several metrics evaluating the 3D geometry are reported: the average Chamfer distance~(CD) to the reference crystal structure linker is of importance as it demonstrates the ability to design linkers of a given shape. The torsion energy~(\(E_{tor}\)) determined with OpenBabel~\cite{openbabel} for the whole molecule is reported. Additionally, a custom shape novelty~(SN) score is introduced, for which the CD to the crystal structure linker (inverse min-max scaled) is multiplied by the Tanimoto diversity (1-similarity) score. The average SN captures our main goal of generating topologically similar, but chemically diverse linkers. In addition, properties of particular relevance to the PROTAC drug modality are reported. These include average number of rings, average number of rotational bonds and fraction of branched linkers. The latter two are properties directly optimized for with ShapeLinker and Link-INVENT and these metrics thus further reflect the optimization capability. To assess chemical plausibility in the context of drug discovery, the average quantitative estimate of drug-likeness~(QED)~\cite{qed_bickerton}, the average SA score~\cite{SA_score} and the fraction passing the 2D filters described in \citet{difflinker} are computed. These include the pan assay interference compounds~(PAINS) filter~\cite{pains} and a ring aromaticity~(RA) filter that ensures rings are either fully aliphatic or aromatic.

\subsubsection{Baselines}
\label{sec:methods_baselines}

Link-INVENT, which is geometry-unconditioned, is used as a baseline. The agent policy of the prior was adapted through RL for every investigated system by including the ratio of rotatable bonds and linker length ratio in the scoring function, but not surface alignment. The two scores are combined in an equally weighted sum. The RL hyperparameters were the same as for ShapeLinker. Additionally, we compare to the pocket-conditioned version of DiffLinker~\cite{difflinker} This method requires specifying the number of atoms making up the new linker, which in our experiments corresponds to the number of atoms found in each reference linker of the crystal structures. The output of DiffLinker is evaluated in two separate ways regarding the geometry: Using the predicted coordinates while allowing replicates of the same constitution (multiple conformers for the same 2D structure) and performing constrained embedding using unique PROTAC SMILES (cf. \ref{sec:const_embed}). The same filters were applied for evaluating the generated poses from the 3D submission as those used for constrained embedding.
\section{Results and Discussion}
\label{sec:results}

\subsection{Shape alignment}
\label{sec:res_align}

The performance of the shape alignment model is assessed by aligning queries to various conformers of themselves and the identical pose from the crystal structure. ShapeLinker can achieve satisfactory results in most instances, though there is variability in performance across the different systems examined (cf. Figure~\ref{fig:shape_val_violin}). This variability can be largely attributed to imperfections in conformer generation, which is also reflected in the RMSD values, with a higher RMSD indicating a larger discrepancy between the generated and the target conformer (cf. Figure~\ref{fig:SI_corr_RMSD_align}). The performance could potentially be enhanced by sampling more conformers, and training on only one reference linker per model, albeit at the expense of increased computational cost. Conformer generation is also the time bottleneck during training of ShapeLinker.

\subsection{Shape conditioning with RL}
\label{sec:res_RL}
\begin{table}[b]
	\centering
	\caption{Chamfer distances between the surface aligned generated extended linkers and the respective crystal structure pose averaged over all systems. To find the best pose, 50~conformers were generated for each system using RDKit.}
	\begin{tabularx}{0.78\textwidth}{ll@{\hskip 0.3in}llll}
    \toprule
        &\multicolumn{4}{c}{\textbf{Chamfer distance}}\\\cmidrule{2-5}
		\textbf{Method}&\textbf{avg $\downarrow$}&  \bm{$<3.5$} \textbf{[\%] $\uparrow$}&\bm{$<2.0$} \textbf{[\%] $\uparrow$}& \bm{$<1.0$} \textbf{[\%] $\uparrow$}\\\midrule
        Link-INVENT&4.44&35.83&8.41&0.18\\
	   ShapeLinker&\bm{$2.19$}&\bm{$88.81$}&\bm{$53.93$}&\bm{$2.9$}\\\bottomrule
	\end{tabularx}
	\label{tab:RL_cmf_ours}
\end{table}
In order to assess the ability of the ShapeLinker models to optimize for shape, samples from trained ShapeLinker were compared to samples taken from Link-INVENT. Table~\ref{tab:RL_cmf_ours} clearly demonstrates this ability as the Chamfer distance between the valid generated samples and the respective crystal structure are lower compared to the geometry-naive model. It should be noted that one could also use the point clouds of pockets instead of the known linker pose for reference-free linker generation. We expect this approach to be most beneficial for cases where the solvent accessible volume available for the linker is restricted by the binding protein(s), leaving a narrow channel that limits potential linker designs. We leave this for further exploration in future studies.

\subsection{Linker generation}
\label{sec:res_linkgen}
\begin{wrapfigure}{r}{0.31\textwidth}
  \begin{center}
    \includegraphics[width=0.29\textwidth]{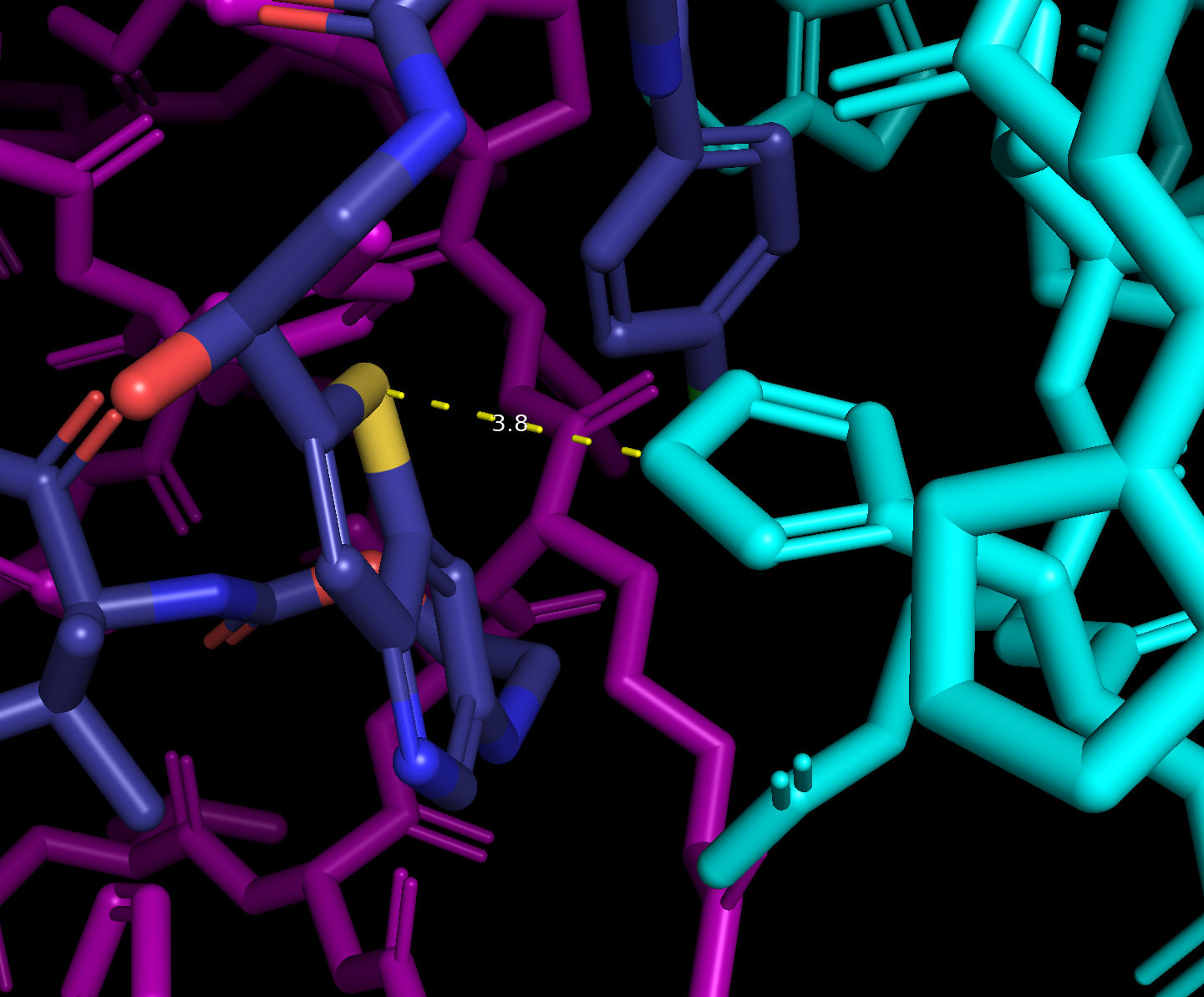}
  \end{center}
  \caption{ShapeLinker-generated linker (dark blue) forming a T-shaped \(\pi\)-stacking between a thiophene moiety and His437 of \(\text{BRD4}^{\text{BD2}}\) (light blue) in 5T35.}
  \label{fig:pi-stack}
\end{wrapfigure}

Two systems, 6BN7 and 6BOY, were excluded from the final analysis as none of the methods performed well on them, which can be expected given the challenging nature of their structure. The reference linkers of these two PROTACs are, together with 7JTO, the longest of the examined systems and also exhibit challenging poses due to the angle between anchor and warhead. The metrics for these two systems are listed in Appendix~\ref{sec:SI_6boy_6bn7}. We argue that this is unlikely to limit practical use significantly, since in a typical drug discovery context one would optimize for less-strained and shorter linkers. \par  

\begin{figure}[b!]
    \centering
    \includegraphics[width=\textwidth]{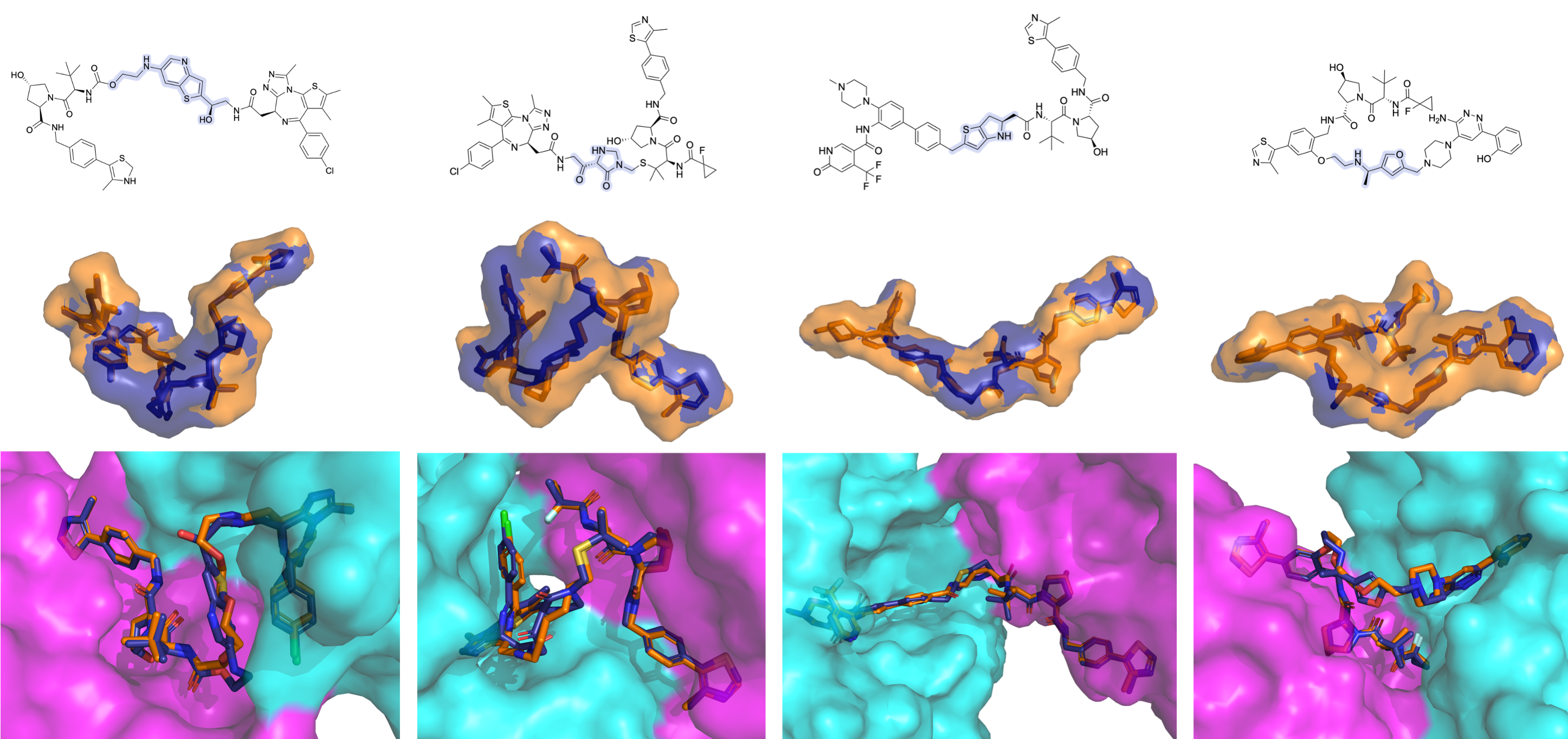}
    \caption{Selected ShapeLinker samples (dark blue) compared to the respective crystal structure PROTAC (orange). \textit{Upper row:} 2D structures with highlighted linker. \textit{Middle row:} aligned surfaces of the reference (orange) and generated PROTAC (blue). \textit{Lower row:} 3D structures binding the E3 ligase (pink) and the POI (light blue). \textit{From left to right:} 5T35, 7ZNT, 7Q2J, 6HAY.}
    \label{fig:example_structures}
\end{figure}

ShapeLinker and Link-INVENT outperform DiffLinker in terms of generative abilities such as validity and uniqueness (see Table~\ref{tab:div_metrics}). The latter is influenced by the choice of number of atoms to generate and providing a range of linker sizes would likely improve uniqueness for DiffLinker. ShapeLinker succeeds in generating very diverse sets of linkers for all systems and further enables to tune the agent by choice of diversity-filter (\textit{vide infra}). Additionally, once trained, sampling is very cheap and the confidence and thus diversity can be modified by varying the temperature scaling of the logits.\par 

\begin{table}[t!]
	\centering
	\caption{Performance metrics evaluating the generative properties of the various methods.}
	\begin{tabularx}{0.7\textwidth}{llll}
    \toprule
		\textbf{Method}&\textbf{Validity [\%]}& \textbf{Uniqueness [\%]}&\textbf{Novelty [\%]}\\\midrule
        Link-INVENT&91.65&\bm{$97.39$}&99.94\\
        DiffLinker&70.81&37.85&99.94\\\midrule
		ShapeLinker&\bm{$93.10$}&95.47&99.94\\\bottomrule
	\end{tabularx}
	\label{tab:div_metrics}
\end{table}

The 3D assessment in Table~\ref{tab:geom_metrics} demonstrates the superiority of the geometry-conditioned ShapeLinker compared to Link-INVENT, further indicating the success of the optimization approach, particularly in terms of improved Chamfer distance. While DiffLinker still achieves lower Chamfer distances, excitingly, our method makes significant progress towards achieving similar Chamfer distances, despite not explicitly sampling coordinates in 3D Euclidean space. In addition, ShapeLinker outperforms DiffLinker in producing the chemical properties required for PROTAC-design, such as producing more rigid linkers with lower number of rotatable bonds, less branching and higher ring count. Notably, also, DiffLinker is limited to a fixed number of atoms, which increases the likelihood of generating viable poses but in turn reduces diversity. Constrained embedding failed for a considerable number of cases for ShapeLinker and Link-INVENT, but not for DiffLinker, where the geometric constraints are already taken into consideration during generation. This is ultimately reflected by DiffLinker's lower validity, which is not only affected by invalid chemistry but also by the failure to actually connect the fragments in space. The goal of generating chemically diverse linkers that are also geometrically similar is embodied in the custom shape novelty (SN) metric. ShapeLinker clearly accomplishes this objective while maintaining strong 2D chemistry metrics. Link-INVENT scores surprisingly well in SN, which is likely attributed to the high number of zero Tanimoto similarity scores (diversity of one). Torsion energies are in general higher than the respective crystal structures (see Table~\ref{tab:case_studies}) for all methods. On one hand, this might be a consequence of attempting to accommodate relatively fixed anchor and warhead poses during constrained embedding, even with subsequent energy minimization. On the other hand, more rigid linkers naturally result in molecules with higher torsional energy compared to reference structures, which predominantly have alkyl chain linkers. It is also worth mentioning that there is great potential in combining the method with tools for ternary complex modelling, alleviating some of the aforementioned uncertainties regarding strained conformations. One could more easily analyse how the new structure might impact the ternary complex but more importantly one could target structures for which there is no crystal structure available.\par 

\begin{table}[b]
    \centering
	\caption{Performance metrics evaluating the ability to generate linkers that lead to molecules with a close geometry to the reference (Chamfer distance (CD)) as well as a good geometry in relation to energetics (torsion energy). The SN score captures the ability to generate linkers with similar shape but new chemistry. \textit{Fail} reports the fraction that failed constrained embedding. \(\text{DiffLinker}_{\text{CE}}\): constrained embedding conformers (deduplicated based on SMILES); \(\text{DiffLinker}_{\text{ori}}\): generated poses with unique conformations but replicate SMILES.}
    \begin{tabularx}{0.68\textwidth}{lllll}
    \toprule
    		\textbf{Method}&\textbf{Failed [\%]} $\downarrow$&\textbf{SN} $\uparrow$&\textbf{CD} $\downarrow$&\bm{$E_{tor}$} $[\frac{\text{kcal}}{\text{mol}}]$ $\downarrow$\\\midrule
        Link-INVENT&27.88&0.82&5.02&69.19\\
        $\text{DiffLinker}_{\text{CE}}$&3.63&0.87&1.96&\bm{$58.24$}\\
        $\text{DiffLinker}_{\text{ori}}$&0.00&0.67&\bm{$1.44$}&60.34\\\midrule
        ShapeLinker&21.45&\bm{$0.9$}&2.64&65.62\\\bottomrule
	\end{tabularx}
	\label{tab:geom_metrics}
\end{table}

Overall, having the 3D context and the rapid ternary screening ability is powerful. For example, one linker generated for 5T35 demonstrated reduced the number of rotatable bonds while introducing a potential new T-shaped \(\pi\)-stacking interaction between a thiophene and His437 of \(\text{BRD4}^{\text{BD2}}\) (see Figure~\ref{fig:pi-stack}). The fused heterocycle in the ShapeLinker-generated linker also significantly rigidifies the structure compared to the PEG-based linker in the reference. Examples shown in Figure~\ref{fig:example_structures} demonstrate the ability to generate linkers adhering to a certain shape. However, they also demonstrate some remaining challenges: all four samples have stereogenic centers complicating synthesis and the example targeting 7ZNT contains a reactive stand-alone carbonyl group negatively impacting stability. Comparable samples generated by Link-INVENT (cf. Figure~\ref{fig:SI_base_example_structures}) do not coincide as well with the reference shape (e.g. example for 7S4E) or clash noticeably with the protein (example for 7JTP). On the other hand, comparable structures produced by DiffLinker (cf. Figure~\ref{fig:SI_difflinker_example_structures}) exhibit similar shape but contain a high number of rotatable bonds. Both samples by Link-INVENT and DiffLinker also exhibit challenges with regard to synthesizability, stability and reactivity. This demonstrates that the domain specific design choices for ShapeLinker yielded significant progress in improved tools for assisting PROTAC-design.\par 

\begin{table}[t]
	\centering
	\caption{Performance metrics assessing the drug-likeness of the generated molecules and the chemical suitability specifically to the class of PROTAC drugs. All metrics focus on the linker fragment only, except for the 2D PAINS filter, which refers to the full PROTAC in order to identify potentially problematic new connections.}
    \begin{tabularx}{\textwidth}{lllllllll}
    \toprule
		\textbf{Method}&\textbf{QED $\uparrow$}&\textbf{SA $\downarrow$}&\textbf{2D Filters [\%] $\uparrow$}&\textbf{\# Rings $\uparrow$}&\textbf{\# ROT $\downarrow$}&\textbf{Branched [\%] $\downarrow$}\\\midrule
		Link-INVENT&\bm{$0.66$}&2.98&92.83&\bm{$1.98$}&3.27&12.06\\
        DiffLinker&0.5&\bm{$2.55$}&\bm{$94.32$}&0.32&2.60&9.66\\\midrule
		ShapeLinker&0.51&3.74&76.51&0.91&\bm{$1.67$}&\bm{$8.64$}\\\bottomrule
	\end{tabularx}
	\label{tab:chem_metrics}
\end{table}

In addition to generating novel designs with a certain shape, ShapeLinker should produce linkers that fit certain 2D criteria for the PROTAC class. Table~\ref{tab:chem_metrics} illustrates that the new method was able to yield linkers with fewer rotatable bonds and little branching. These results, together with the challenging task of matching the query shape, were achieved at the cost of QED, SA and relatedly, the ratio that passed the 2D filters. A more permissive choice of diversity filter could also help with improving QED and SA of ShapeLinker, as there would be less incentive to steer away from the prior distribution trained on the drug-like chemical space. The baseline model was optimized for branching and ratio of rotatable bonds but still failed to outperform DiffLinker. The fact that Link-INVENT achieves the best result for the number of rings is likely attributable to the absence of linker size limitations, which often leads to the generation of excessively long designs. The inclusion of the number of rings as a score for ShapeLinker or Link-INVENT is possible if an increase in this metric is desired. The combined results demonstrate the inability of Link-INVENT to generate linkers fitting a desired shape while DiffLinker lacks diversity. ShapeLinker addresses these limitations and combines favorable aspects of both.

\section{Conclusion}
\label{sec:conclusion}
This work introduces a novel method, ShapeLinker, for generating novel PROTAC linkers adhering to a target conformation. It introduces a highly modular and expandable Reinforcement Learning-framework to specifically address limitations of existing works in the optimization of PROTAC-linkers. In addition to performing well across existing linker generation related metrics, it achieves excellent performance in shape novelty, which captures a models ability to generate novel chemical matter that can assume a desired shape. It further illustrated the successful combination of well-established multi-parameter optimization techniques for autoregressive models with a novel shape alignment approach for additional scoring. ShapeLinker enables \textit{de novo} design of linker fragments suited to the PROTAC drug modality especially during lead optimization. Despite generating in 2D and having no geometry input at inference time, it approaches the performance of fully 3D aware methods such as DiffLinker while enabling flexible and modular combination of several, program-specific composite scoring functions which are not as easily incorporated in 3D diffusion-based methods like DiffLinker. We demonstrate this by showing the optimization towards simple physicochemical constraints, a valuable property for PROTAC molecules, which typically fall far beyond the "rule of 5". Additionally, once the augmented agent is trained, the sampling with ShapeLinker has minimal computational cost and inference time.\par 

Possible extensions of ShapeLinker include incorporating the QED and SA metrics directly into the multi-parameter optimization or opting for a more permissive or no-diversity filter to better leverage the learned semantics of the pre-trained model. A future endeavor should also be the inclusion of biopharmaceutically relevant scores such as predictors for solubility or even phenotypic degradation effect. Lastly, the use of the pocket shape for alignment instead of a reference conformer is worth investigating and could open new avenues to explore.


\begin{ack}
The authors thank Andrew~G.~Tsesis for helping with experiments using DiffLinker. The authors also thank Dylan~Abramson, Jeff~Guo, Ilia~Igashov, Haichan~Niu, Chalada~Suebsuwong and Xuejin~Zhang for helpful suggestions regarding the structuring and content of this paper.
\end{ack}

\section*{Conflicts of interest}
RN, MA, DK, LN were employed by VantAI during the time of writing

\bibliographystyle{unsrtnat}
\bibliography{ref}  
\newpage
\appendix

\setcounter{section}{0}
\setcounter{table}{0}
\setcounter{figure}{0}
\renewcommand\thesection{S\arabic{section}}
\renewcommand\thesubsection{\thesection.\arabic{subsection}}
\renewcommand{\thefigure}{\thesection.\arabic{figure}}
\renewcommand{\thetable}{\thesection.\arabic{table}}

\section{Shape alignment}
\label{sec:SI_align}

\subsection{Point cloud generation}
\label{sec:SI_pcgen}

We adapt the surface point cloud generation procedure delineated in dMaSIF, which samples the molecular surface of a protein as a level set of the smooth distance function to atom centers. The sampling algorithm first generates a point cloud in the neighborhood of a protein and then lets the random sample converge towards the target level set via gradient descent. Subsequently, points trapped inside the protein are removed, ensuring uniform density by averaging samples within cubic bins of side length 1~\si{\angstrom}. However, this procedure is designed for protein surfaces, which is not the focus of our use case.

To adapt the procedure for small molecule surfaces, we modify the method by reducing the radius to 0.9~\si{\angstrom} and decreasing the resolution to 0.9~\si{\angstrom}, thereby achieving a higher density of points on the surface. Additionally, we introduce an "other" atom type to encompass all types that are not defined in dMaSIF (defined: C, H, O, N, S, Se). For this "other" atom type, we use the radius of a carbon atom. These alterations accommodate the smaller size and distinct characteristics of small molecule surfaces compared to protein surfaces.

\subsection{Architecture details}
\label{sec:SI_shape_arch}

In this section, we outline the process by which both query surface points ($n$ 3-dimensional coordinates) and reference surface points ($m$ points) are transformed using attention layers to generate pseudo-coordinates and, ultimately, aligned-coordinates of the query molecule. Refer to Figure~2 for a visual representation.

\subsubsection{Self-attention encoder}

Initially, the reference (\(\mathbf{R}\)) and query (\(\mathbf{Q}\)) points are centered at the origin. They are then passed through a fully connected linear layer (\(\mathbf{Linear}\)) to scale them to the attention embedding dimension \(d_a\) of 16:
\begin{equation}
\mathbf{Q}_{scaled} = \mathbf{Linear}(\mathbf{Q}, d_a), \quad \mathbf{R}_{scaled} = \mathbf{Linear}(\mathbf{R}, d_a)
\end{equation}
The scaled query and reference points are subsequently processed through the same self-attention network (\(\mathbf{SelfAttention}\)), characterized by an attention dimension of 16 and an attention head size \(h\) of 8. For both the query and reference, the query (\(q\)), key (\(k\)), and value (\(v\)) are the scaled inputs (\(\mathbf{Q}_{scaled}\) and \(\mathbf{R}_{scaled}\)).

\begin{equation}
\mathbf{Q}_{self\_attention} = \mathbf{SelfAttention}(q_\mathbf{Q}, k_\mathbf{Q}, v_\mathbf{Q}, d_a, h)
\end{equation}
\begin{equation}
\mathbf{R}_{self\_attention} = \mathbf{SelfAttention}(q_\mathbf{R}, k_\mathbf{R}, v_\mathbf{R}, d_a, h)
\end{equation}

\subsubsection{Cross-attention decoder}

Afterwards, the output from the query self-attention serves as the query, while the reference output is used as both keys and values in the cross-attention network (\(\mathbf{CrossAttention}\)), which shares the same attention and head size as the self-attention network:

\begin{equation}
\mathbf{Q}_{cross\_attention} = \mathbf{CrossAttention}(q_\mathbf{Q}, k_\mathbf{R}, v_\mathbf{R}, d_a, h)
\end{equation}

Consequently, the output from the cross-attention network is scaled using a dense linear layer to a dimension \(d_o\) of 3, representing the pseudo-coordinates:

\begin{equation}
\mathbf{Q}_{pseudo} = \mathbf{Linear}(\mathbf{Q}_{cross\_attention}, d_o)
\end{equation}
Finally, the Kabsch algorithm is applied to superimpose the original query input onto the pseudo-coordinates, resulting in the aligned-coordinates of the query. The aligned-coordinates, along with the reference coordinates, are subjected to the L2 normalized chamfer loss (defined in Section~3.1):

\begin{equation}
\mathbf{Q}_{aligned} = \mathbf{Kabsch}(\mathbf{Q}_{pseudo}, \mathbf{R})
\end{equation}

\subsubsection{Alignment inference}

In the alignment inference, there are two modes:
\begin{enumerate}
    \item The first mode involves returning the surface point clouds of the query, accompanied by the chamfer distance to the reference point cloud, which can be fed into the RL-training process.
    \item Since the alignment process yields the rotation and translation matrices, these can be utilized post-training to transform the original atom coordinates of a given query.
\end{enumerate}

\subsection{Caveats}
Using only the extended linker fragment can result in relatively linear fragments to align, which may cause the model to align the poses in a flipped orientation. To avoid this, the shape alignment is repeated for those with high RMSD of substructure matches. During RL, resampling is performed until 90\% of the samples have an RMSD matching the lower distribution, or for a maximum of 5 iterations. The shape alignment carried out during post-processing is done exhaustively, ensuring that all samples have a fitting alignment.

\begin{figure}[b]
    \begin{center}
    \includegraphics[width=0.5\textwidth]{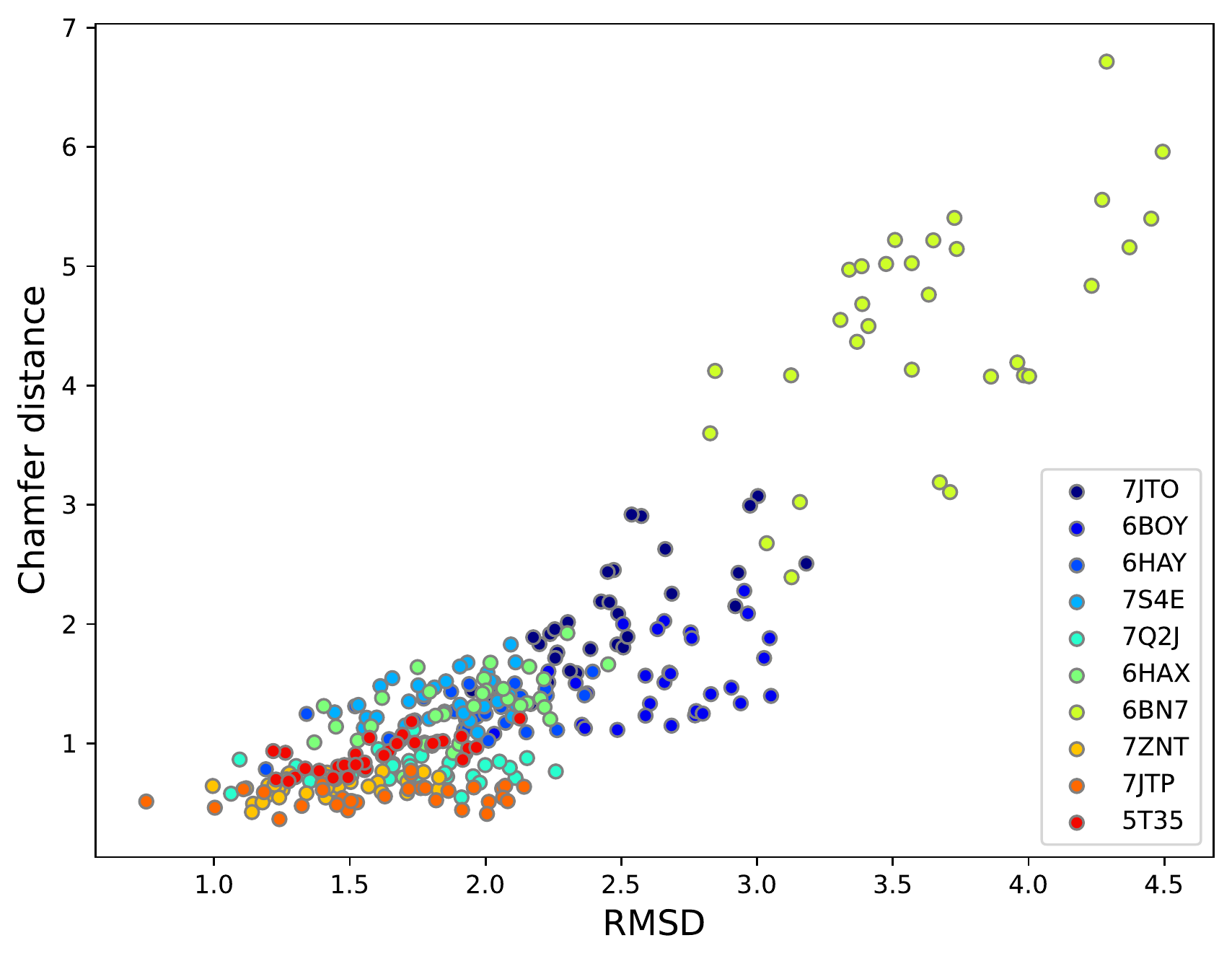}
    \caption{Correlation of Chamfer distance to RMSD obtained by the shape alignment model. The randomly generated conformers are compared to the pose found in the corresponding crystal structure.}
    \label{fig:SI_corr_RMSD_align}
    \end{center}
\end{figure}

\subsection{Performance}
\label{sec:SI_align_performance}
\begin{figure}[h]
    \begin{center}
    \includegraphics[width=0.5\textwidth]{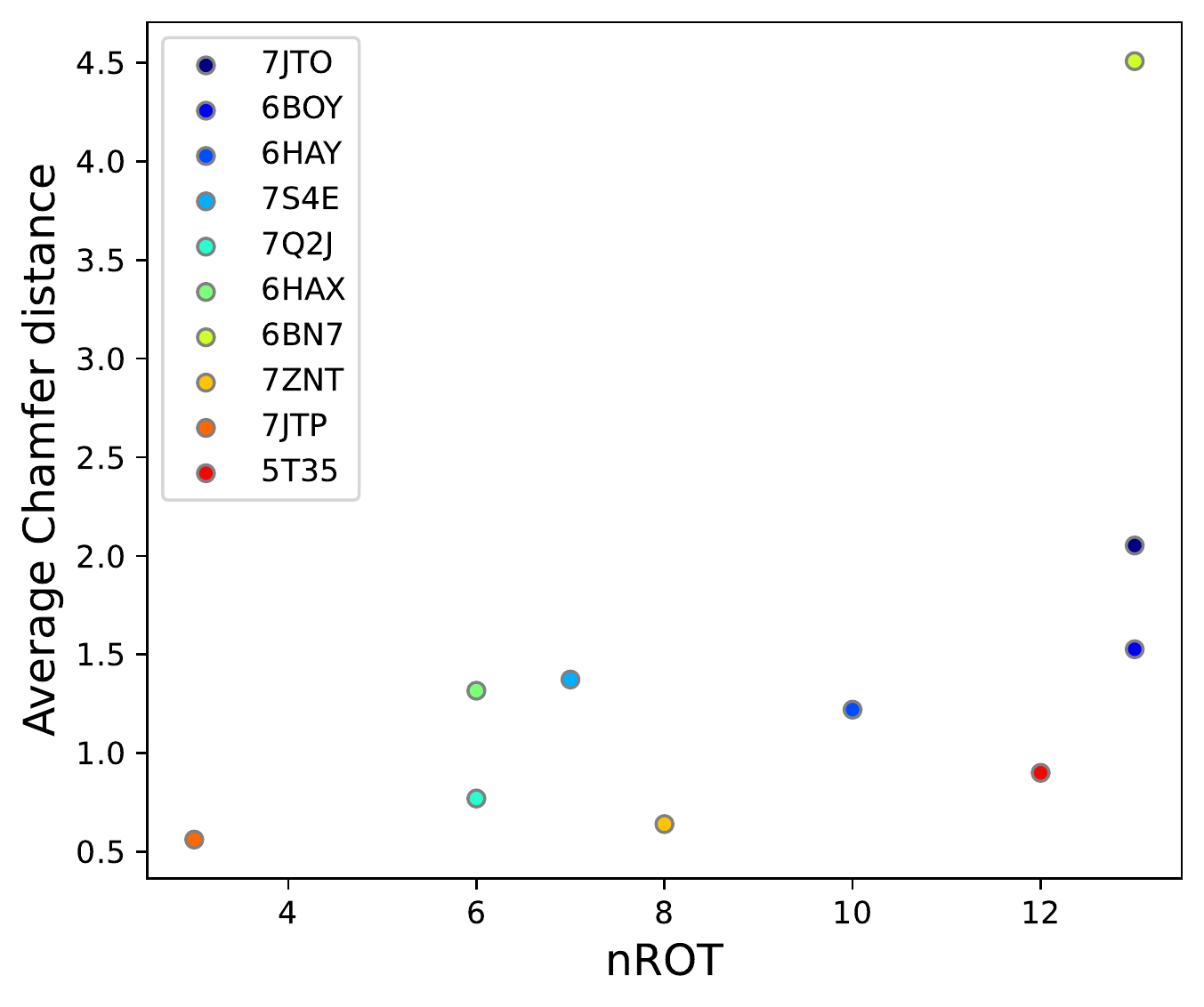}
    \caption{Correlation of the average chamfer distance ($n = 32$) obtained by the shape alignment model to the number of rotational bonds. The randomly generated conformers are compared to the pose found in the corresponding crystal structure.}
    \label{fig:SI_corr_nROT_align}
    \end{center}
\end{figure}

\begin{figure}[h]
    \begin{center}
    \includegraphics[width=0.7\textwidth]{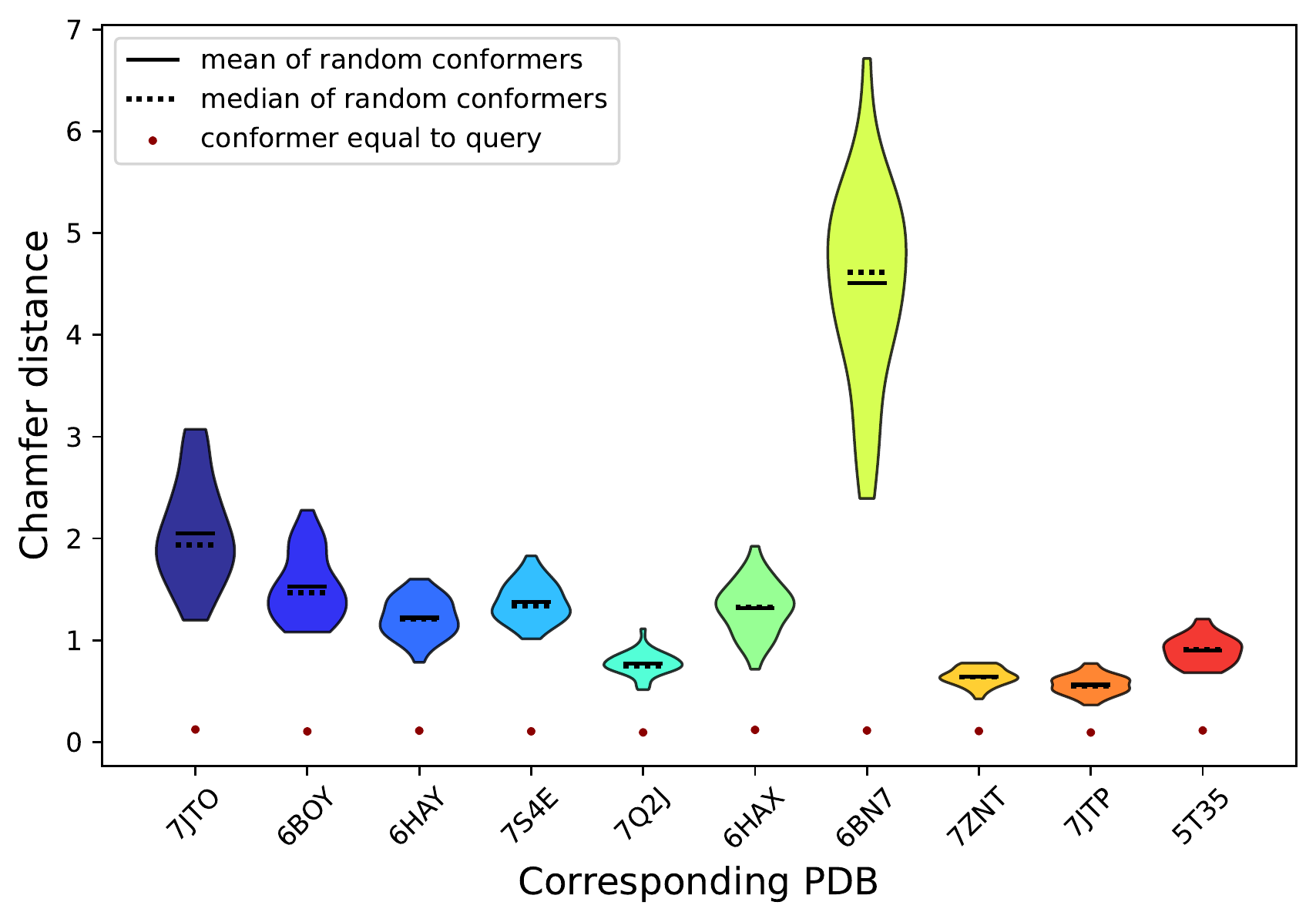}
    \caption{Distribution of Chamfer distances per structure obtained by aligning each extended linker (cf. Figure~\ref{fig:SI_all_ref_strucs}) to the pose found in the crystal structure (32 distances obtained by aligning 16~conformers each). The red dot corresponds to the Chamfer distance obtained when comparing the surfaces of the identical poses. Good alignment is expected in the range of each Chamfer distance distribution for the respective system while the red dot corresponds to the best score possible.}
    \label{fig:shape_val_violin}
    \end{center}
\end{figure}

The Chamfer distance resulting from the alignments are correlated with the number of rotational bonds (Pearson's r of 0.55) and RMSD (Pearson's r of 0.88) (see Figures~\ref{fig:SI_corr_nROT_align} and \ref{fig:SI_corr_RMSD_align}), which further supports the notion that the main source of variability in the performance of the shape alignment model is the conformer generation rather than the alignment process itself.
A good alignment for a specific pose is determined by the upper to lower bounds of the Chamfer distance, as demonstrated in Figure~\ref{fig:shape_val_violin}. The impact of a high number of degrees of freedom on performance is counterbalanced in the design of ShapeLinker's multi-parameter optimization, where the model is trained to generate linkers with fewer rotational bonds, resulting in more rigid structures.

\clearpage
\newpage

\section{Training of the Link-INVENT based methods}
\label{sec:SI_RL}
Both Link-INVENT and ShapeLinker were trained (RL) for 1000~epochs with a batch size of 32 and a learning rate of 1e-4. Both methods apply a diversity filter as implemented in Link-INVENT. All sampled molecules during RL are collected in "buckets" sharing the same Murcko scaffold. If the bucket reaches 25~samples, all subsequently generated molecules with the same scaffold will be penalized with a score of zero -- thereby urging the model to explore a new chemical space. 5,000~linkers are sampled using a temperature scaling of 1.5 for for each examined system for both methods. The models were trained using one~GPU (NVIDIA T4) and eight~CPU cores (Intel Broadwell) on the Google Cloud Platform.

\subsection{Baseline Link-INVENT}
\label{sec:SI_RL_base}

\begin{figure}[htbp]
\centering
\subfigure[Average score.]{\includegraphics[width=0.47\textwidth]{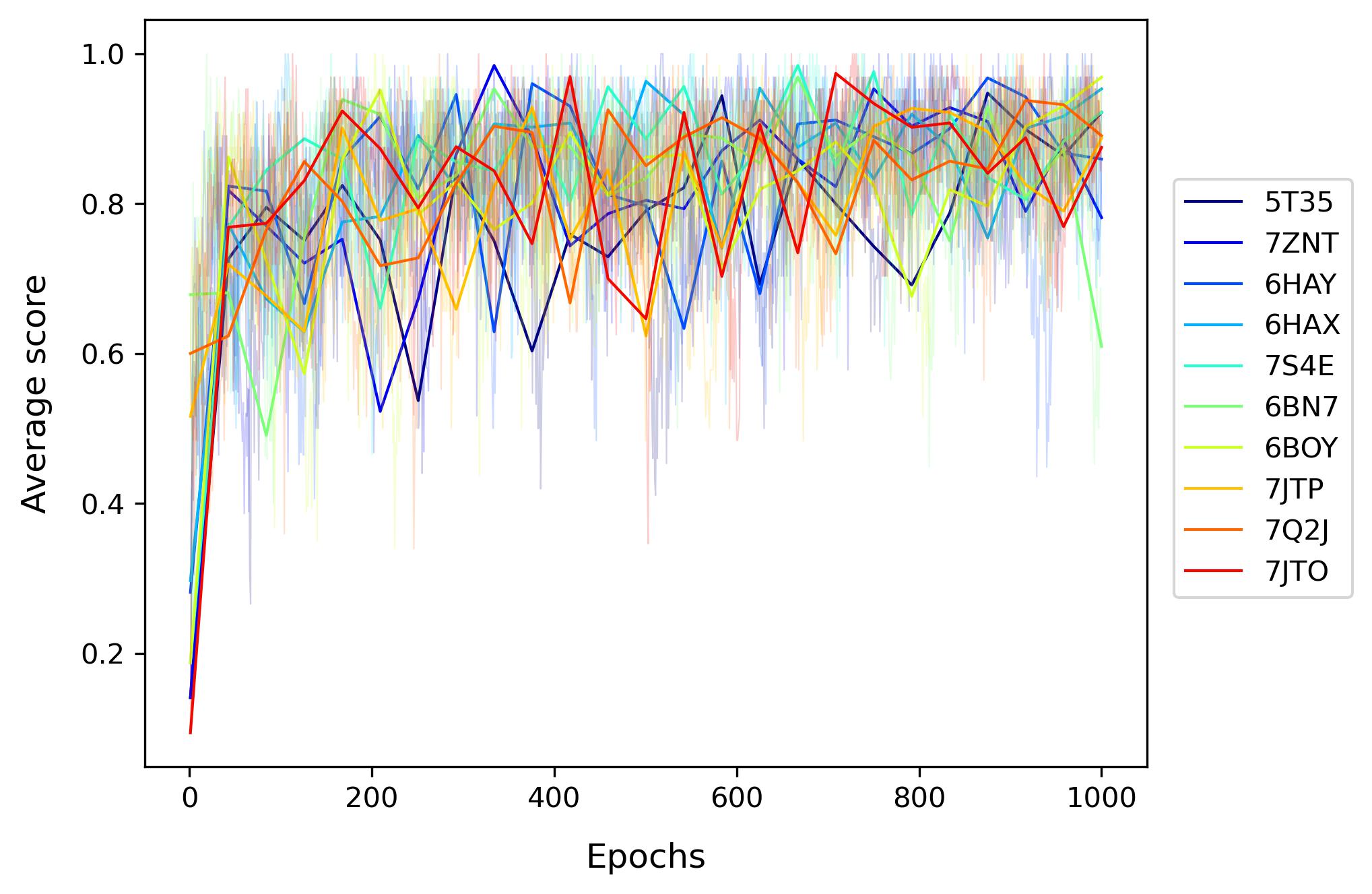}}\label{subfig:SI_base_lc_avg}
\hfill
\subfigure[Transformed linker length ratio.] {\includegraphics[width=0.47\textwidth]{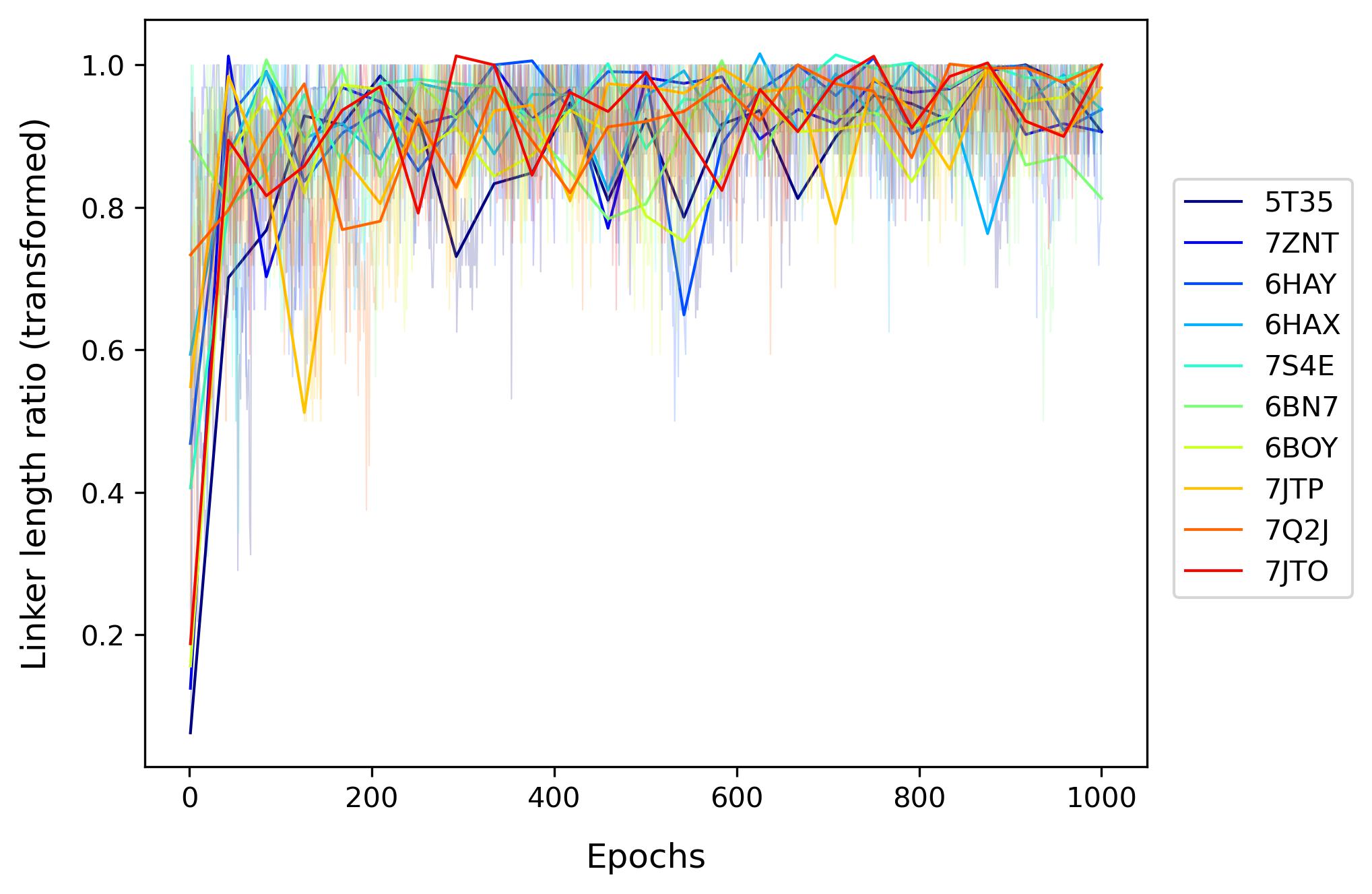}}\label{subfig:SI_base_lc_len}
\subfigure[Transformed ratio of rotatable bonds.]{\includegraphics[width=0.47\textwidth]{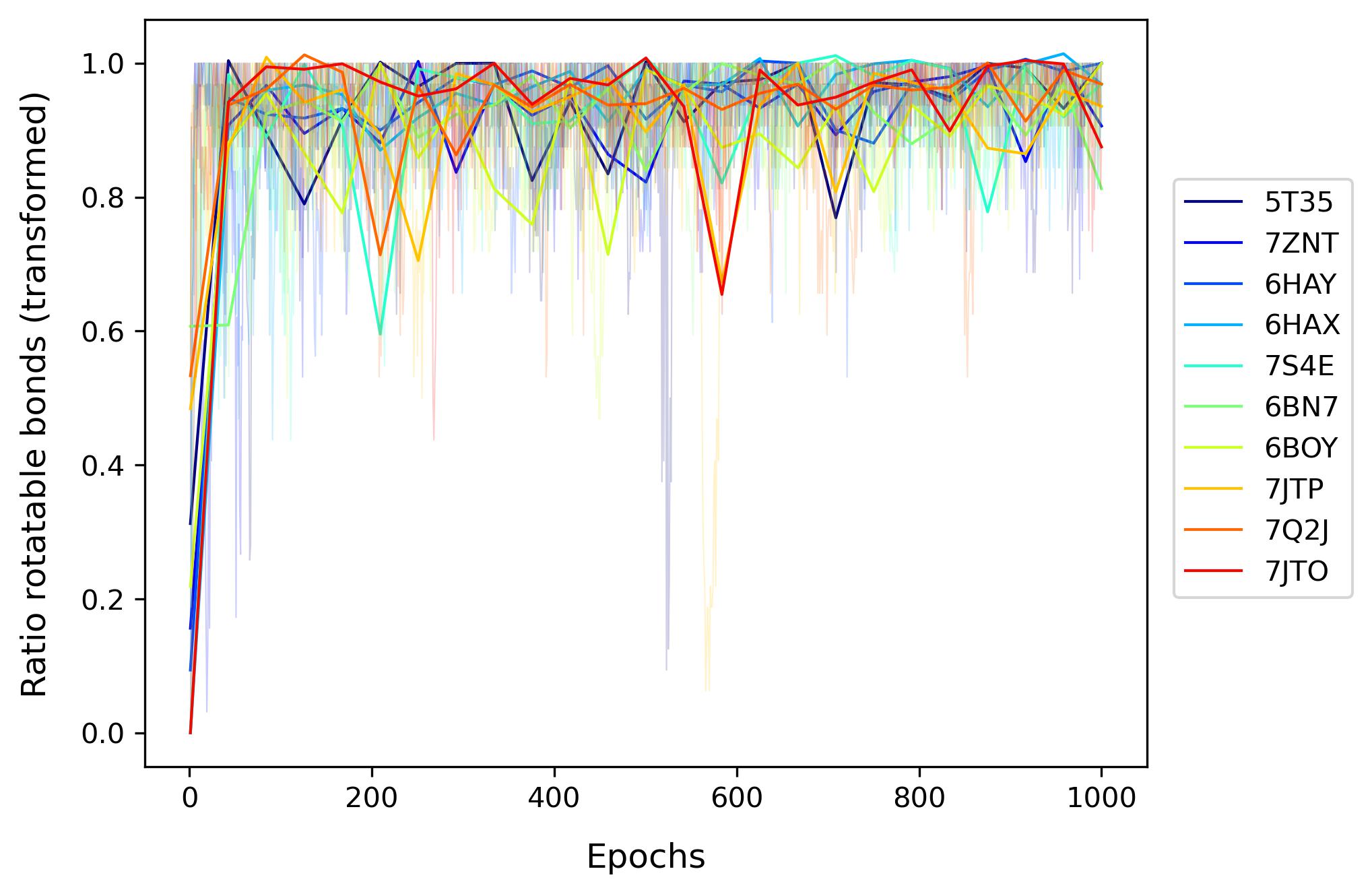}}\label{subfig:SI_base_lc_nrot}
\caption{Learning curves for all baseline Link-INVENT RL runs. The average score combines the linker length ratio, ratio of rotatable bonds and factors in the penalty by the diversity filter.} \label{fig:SI_base_lc}
\end{figure}

\subsection{ShapeLinker: Geometry-conditioned Link-INVENT} 
\label{sec:SI_RL_ours}

\begin{figure}[htbp]
\centering
\subfigure[Average score.]{\includegraphics[width=0.47\textwidth]{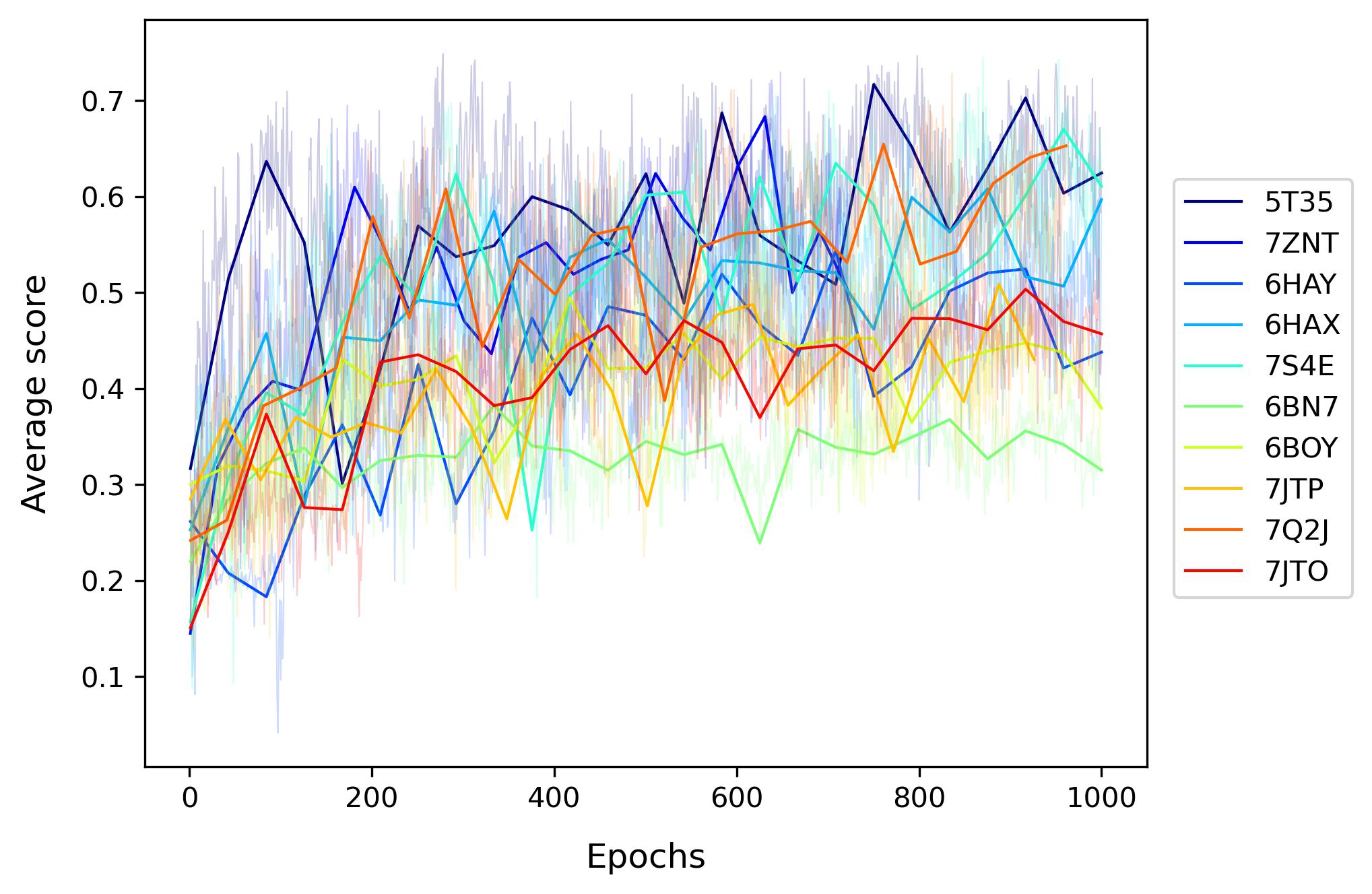}}\label{subfig:SI_shape_lc_avg}
\hfill
\subfigure[Transformed linker length ratio.] {\includegraphics[width=0.47\textwidth]{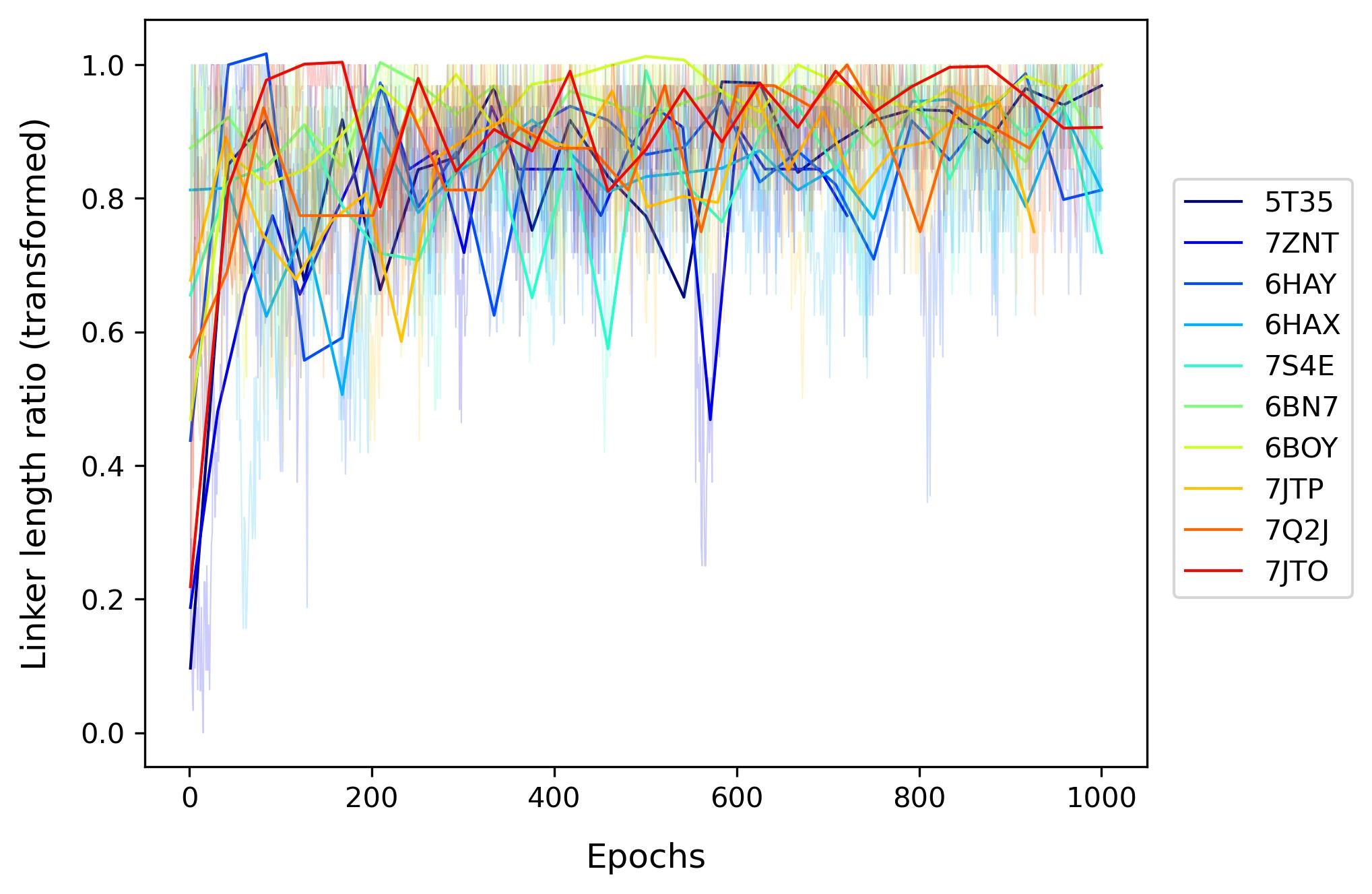}}\label{subfig:SI_shape_lc_len}
\subfigure[Transformed ratio of rotatable bonds.]{\includegraphics[width=0.47\textwidth]{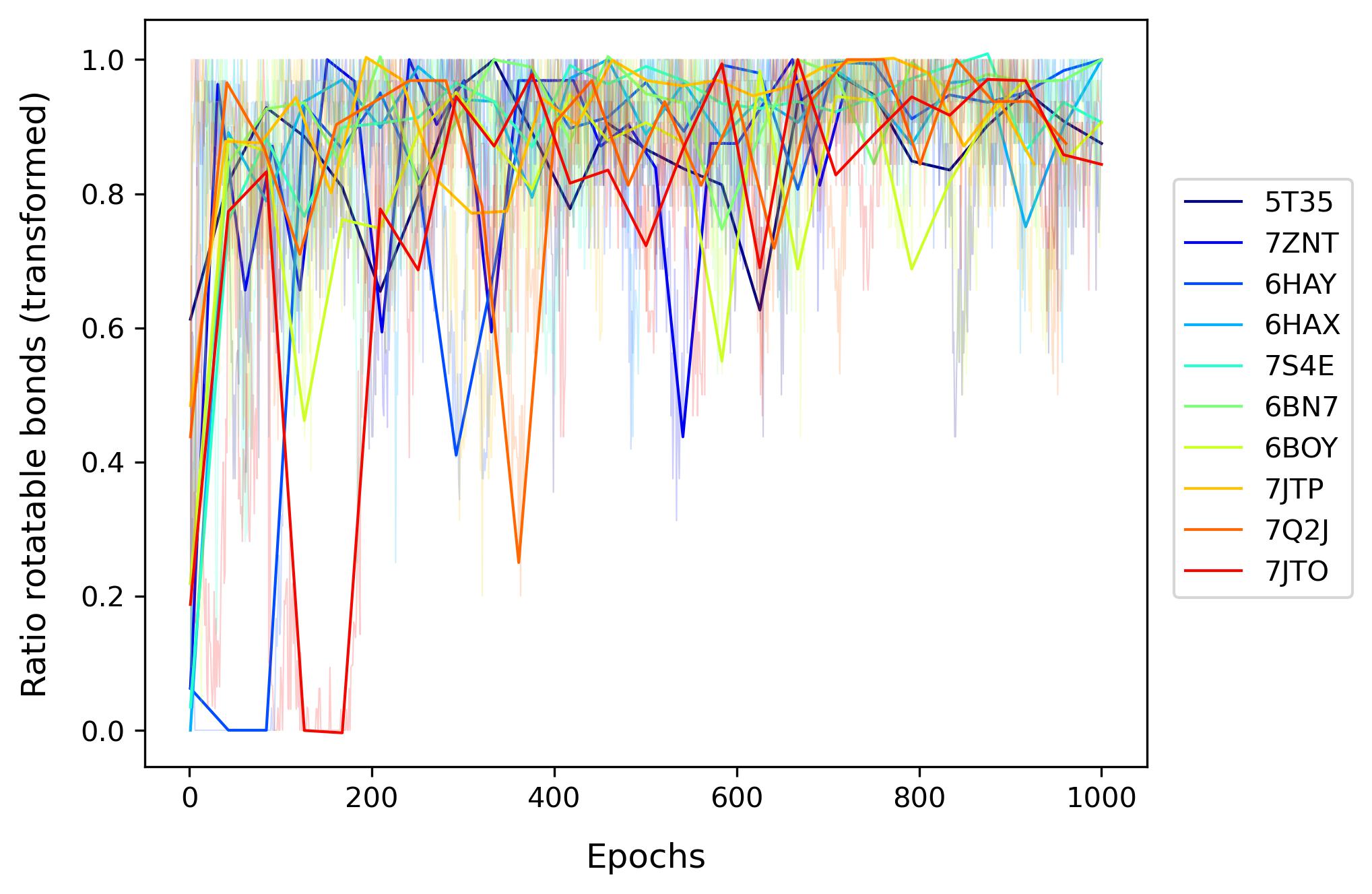}}\label{subfig:SI_shape_lc_nrot}
\hfill
\subfigure[Transformed chamfer distance.]{\includegraphics[width=0.47\textwidth]{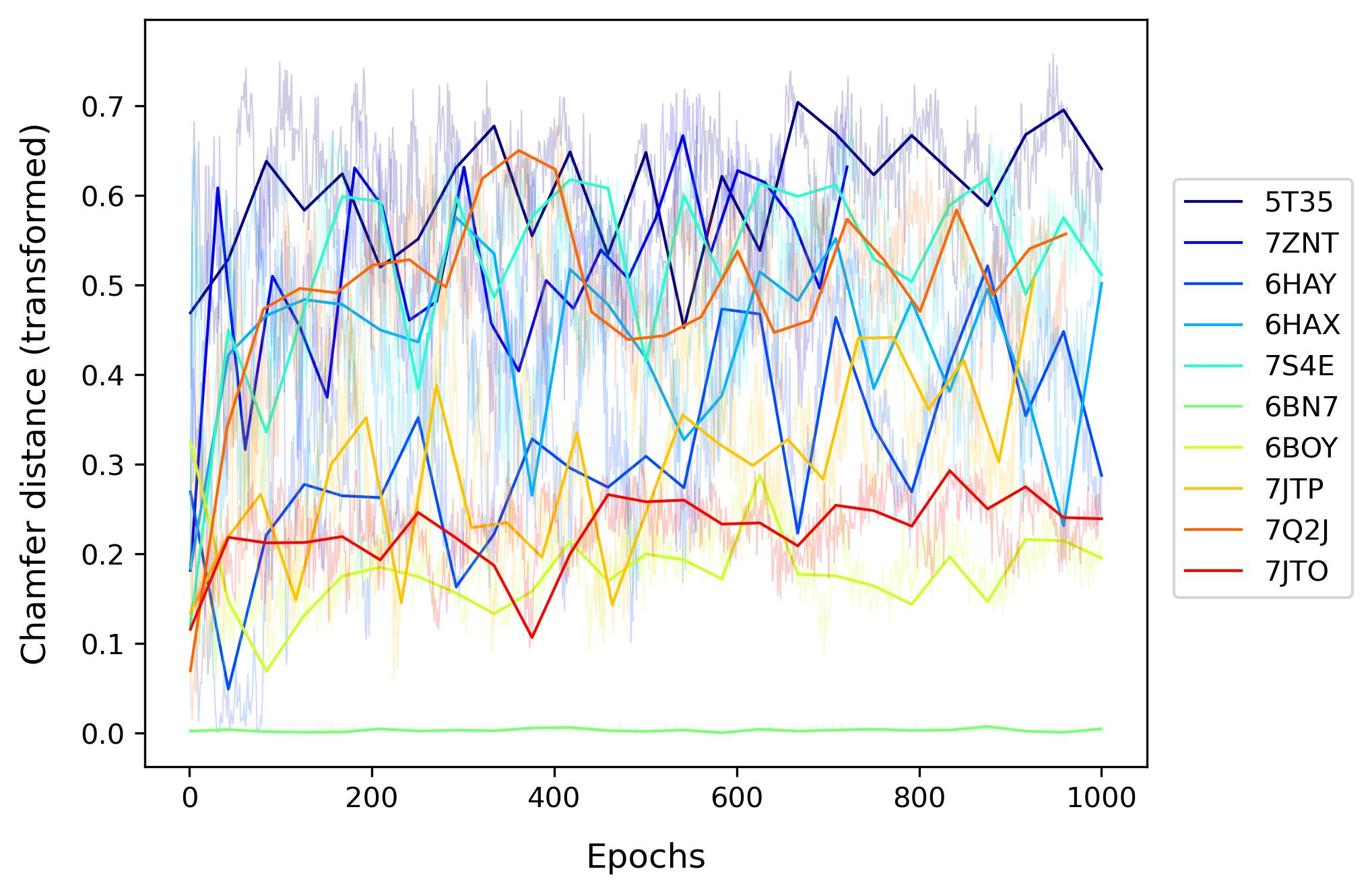}}\label{subfig:SI_shape_lc_shape}
\caption{Learning curves for all ShapeLinker RL runs. The average score combines the linker length ratio, ratio of rotatable bonds, shape alignment score and factors in the penalty by the diversity filter.} \label{fig:SI_shape_lc}
\end{figure}

The alignment during RL is carried out on the level of the extended linker with 16~conformers generated for each linker and the smallest distance of those corresponds to the raw score for sample~$x$. All models were intended to train for 1,000 epochs each, but 7ZNT (720~epochs), 7JTP (920~epochs), and 7Q2J (960~epochs) were interrupted early due to unknown reasons. Since all three models had already converged for all objectives, the last logged agent was used for subsequent sampling. The learning curves for the shape alignment (see Figure~\ref{subfig:SI_shape_lc_shape}) are quite noisy. In addition to the challenging task of learning a 3D objective while generating SMILES, this is likely due to the shape alignment model's inability to correctly process charged structures. In such cases, scores of zero are automatically returned. Both the baseline Link-INVENT and ShapeLinker could converge towards low number of rotatable bonds and low linker length ratio early during training. \par 
Given the early convergence for most systems, one could likely sample from earlier epochs (where the average score has already converged) and expect a different chemical space as a result of the diversity filter steering the generation towards novel chemistry.

\section{Data}
\label{sec:SI_data}

\subsection{PROTAC-DB}
\label{sec:SI_protacdb}
A total of 3,270 SMILES for PROTAC, anchor, and warhead each were extracted from PROTAC-DB.\cite{protacdb} Forty-seven faulty entries, which had no substructure match for either the warhead or anchor to the PROTAC, were removed. Additionally, 41 instances were excluded due to unsuccessful extraction of the extended linker fragment, which contains the linker as well as fragments extending beyond the exit vector. The extraction of the extended linkers was carried out in such a way as to preserve the geometry of the bonds between the linker and the respective POI and E3 ligands, the extended linker is extracted at least two hops from the attachment point, while ensuring that no rings are broken and bond order is not changed. The removed PROTAC-DB IDs are as follows:
\begin{quote}
67,
 90,
 164,
 632,
 633,
 634,
 635,
 636,
 637,
 638,
 639,
 640,
 641,
 1001,
 1032,
 1047,
 1049,
 1060,
 1153,
 1198,
 1302,
 1303,
 1535,
 1536,
 1949,
 1950,
 1951,
 1952,
 1953,
 1954,
 1955,
 1956,
 1957,
 1958,
 1959,
 1960,
 1961,
 1962,
 1963,
 1964,
 1965,
 1966,
 1967,
 1968,
 1969,
 2110,
 2196,
 2237,
 2238,
 2239,
 2240,
 2241,
 2242,
 2243,
 2244,
 2245,
 2246,
 2276,
 2381,
 2382,
 2383,
 2384,
 2385,
 2386,
 2387,
 2388,
 2389,
 2390,
 2442,
 2443,
 2529,
 2545,
 2876,
 2959,
 2962,
 2966,
 2967,
 2968,
 3129,
 3213,
 3214,
 3215,
 3216,
 3217,
 3218,
 3219,
 3220,
 3221
 \end{quote}
 For the training of the shape alignment model, the extended linker poses of all 10 investigated crystal structures were used as queries. The training set consisted of 50~conformations of each query structure to learn self alignment and 50~extended linkers each randomly selected from the processed PROTAC-DB dataset to learn alignment to other structures. The validation set consisted of 10~extended linkers from the PROTAC-DB dataset. All conformations were generated using RDKit with random coordinates.

 \subsection{Crystal structures of the investigated ternary complexes}
 \label{sec:SI_xtal_data}

\begin{figure}[h]
    \centering
    \includegraphics[width=\textwidth]{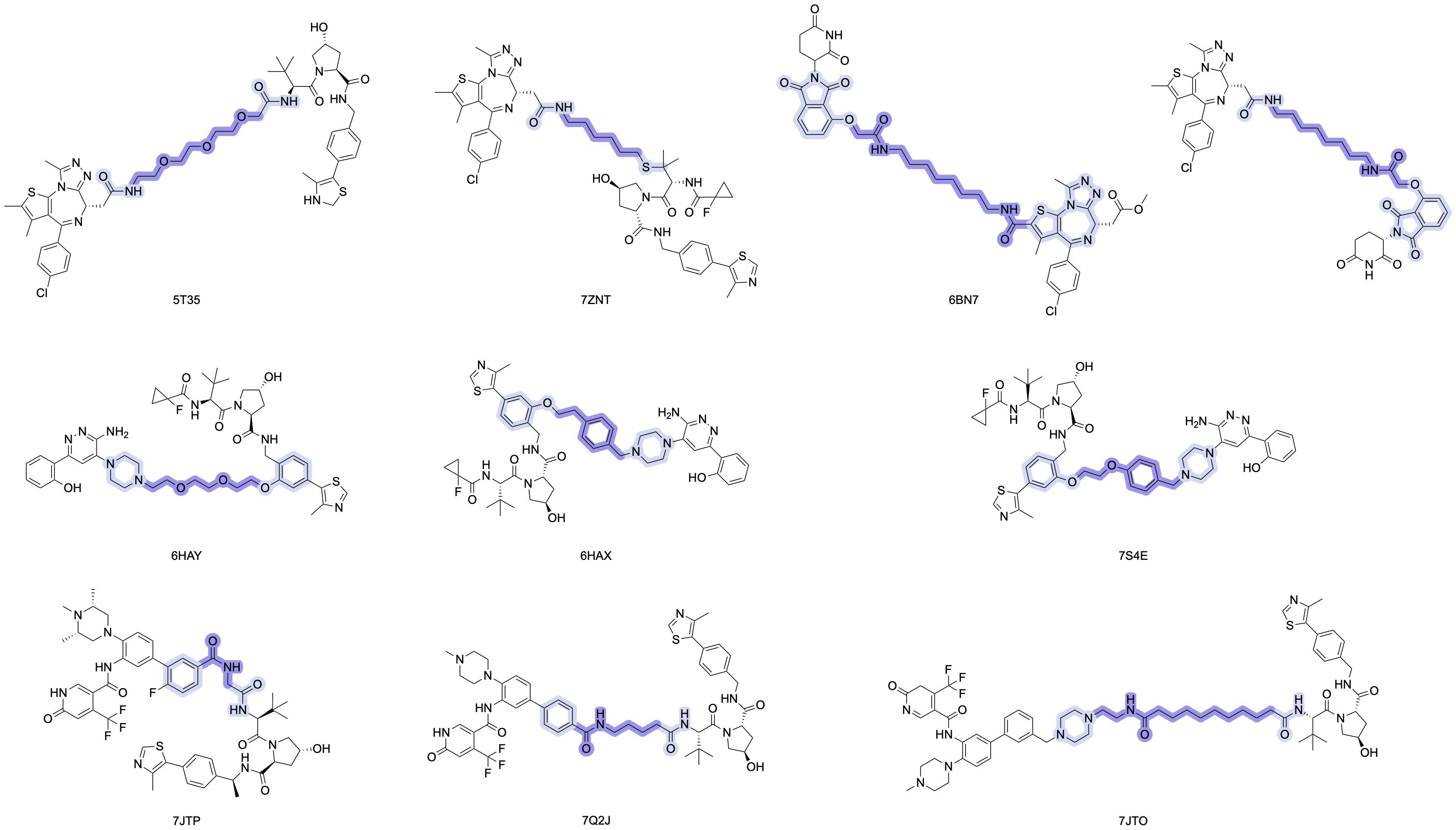}
    \caption{Chemical structures of all reference PROTACs binding the investigated crystal structures. Highlighted in dark blue is the linker, which was cut out for generation of new linkers and highlighted in light blue are the additional fragments for the extended linkers, which were used for the shape alignment.}
    \label{fig:SI_all_ref_strucs}
\end{figure}

 The crystal structures were prepared by extracting one asymmetric sub-unit of the ternary complex and removing any solvents or crystallization artifacts.  The selected PROTACs cover a diverse range of shapes and lengths. The linker fragment was selected in accordance with the authors of the structures, with the constraint of keeping the flanking amide bonds intact (either belonging to the linker or the anchor/warhead). This approach was taken in hopes of reducing the risk of generating synthetically challenging termini. The chosen fragmentation for all investigated systems can be seen in Figure~\ref{fig:SI_all_ref_strucs}.

\begin{table}[H]
\small
	\centering
	\caption{Chosen systems with the respective targeted protein of interest (POI) and E3 ligase, the PDB~ID for the crystal structure of the ternary complex and lastly the name of the reference PROTAC. Calculated torsion energy and number of clashes for each conformation of the PROTAC are listed.}
	\begin{tabularx}{0.77\textwidth}{llllll}
    \toprule
		\textbf{POI}&\textbf{E3}& \textbf{PDB ID}&\textbf{PROTAC}&\bm{$E_{tor}$} $[\frac{\text{kcal}}{\text{mol}}]$ $\downarrow$&\textbf{\# Clashes $\downarrow$}\\\midrule
        \multirow{2}{*}[-0.09cm]{$\text{BRD4}^{\text{BD2}}$}&\multirow{2}{*}[-0.09cm]{VHL}&5T35 \cite{5T35_xtal}&MZ1&63.86&10\\\cmidrule{3-6}
		&&7ZNT \cite{7ZNT_xtal}&AT7&41.74&10\\\midrule
        \multirow{2}{*}[-0.09cm]{$\text{BRD4}^{\text{BD1}}$}&\multirow{2}{*}[-0.09cm]{CRBN}&6BN7 \cite{6BN7_6BOY_xtal}&dBET23&33.67&20\\\cmidrule{3-6}
		&&6BOY \cite{6BN7_6BOY_xtal}&dBET6&25.56&10\\\midrule
        \multirow{3}{*}[-0.2cm]{SMARCA2}&\multirow{3}{*}[-0.2cm]{VHL}&6HAY \cite{6HAX_6HAY_7S4E_xtal}&PROTAC 1&50.20&11\\\cmidrule{3-6}
        &&6HAX \cite{6HAX_6HAY_7S4E_xtal}&PROTAC 2&43.58&6\\\cmidrule{3-6}
		&&7S4E \cite{6HAX_6HAY_7S4E_xtal}&ACBI1&37.86&15\\\midrule
        \multirow{3}{*}[-0.2cm]{WDR5}&\multirow{3}{*}[-0.2cm]{VHL}&7JTP \cite{7JTO_7JTP_xtal}&MS67&56.07&16\\\cmidrule{3-6}
        &&7Q2J \cite{7Q2J_xtal}&-&55.53&21\\\cmidrule{3-6}
		&&7JTO \cite{7JTO_7JTP_xtal}&MS33&42&18\\\midrule
	\end{tabularx}
	\label{tab:case_studies}
\end{table}

\section{Constrained embedding}
\label{sec:SI_const_embed}

\begin{figure}[h]
    \centering
    \includegraphics[width=0.45\textwidth]{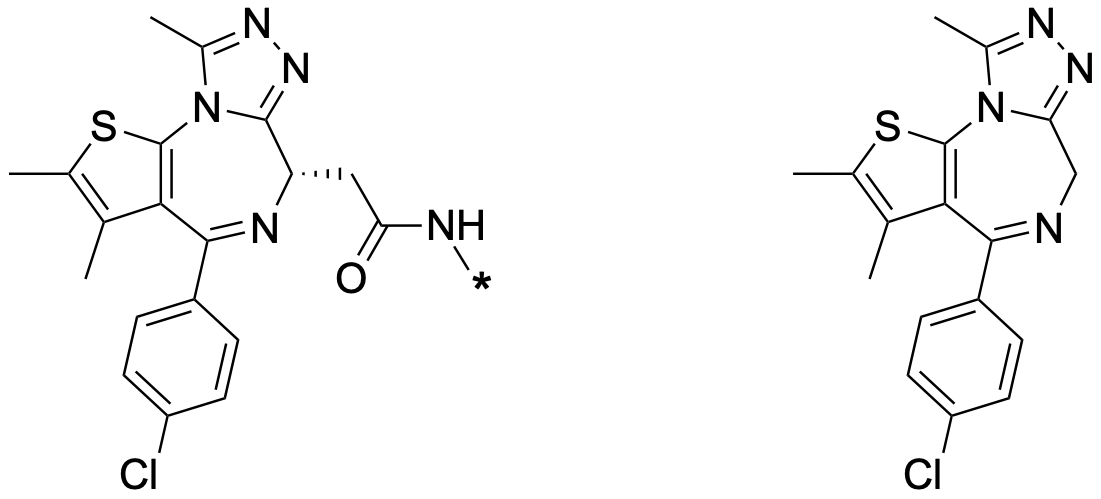}
    \caption{Structure of the BRD4 warheads showing the modification that was required for constrained embedding. Left: Complete warhead used for generation. Right: Reduced warhead used for constrained embedding.}
    \label{fig:SI_brd4_warhead}
\end{figure}

The preparation of samples for constrained embedding from both Link-INVENT-based methods required annotation of stereocenters, including chiral centers and \textit{cis}/\textit{trans} bonds. This was achieved by shape aligning all samples to the crystal structure pose. To capture potential stereocenters at the exit vector (the bond between the attachment atoms of the linker and anchor/warhead), the extended linker and the same shape alignment model used during RL were used. The stereocenters for DiffLinker could be directly annotated from the generated pose. In case RDKit fails to annotate some stereocenters (e.g. some bridge heads), the isomers will be enumerated and all submitted to the constrained embedding. The same fragments for anchor and warhead were used as constraints during the embedding with the exception of BRD4-binding warheads (5T35, 7ZNT, 6BN7, 6BOY), where there were no productive poses found for the Link-INVENT based methods. The warhead fragment used to constrain the embedding was reduced by removing the flexible chain that includes the exit atom and is attached to the core ring (see Figure~\ref{fig:SI_brd4_warhead}). This alteration should not introduce bias, since the chain is flexible and can move during minimization, and 3D evaluation is ultimately done on the whole warhead. Despite the modification, 6BN7 and 6BOY still did not result in any productive poses and their challenging nature was discussed in the main text.\par

Initially, the generation of 10~conformers each with constrained embedding was attempted using RDKit~\cite{rdkit}, allowing a maximum of 10~attempts. For SMILES that did not result in a productive pose, this process was repeated by increasing the maximum attempts up to 10,000 while decreasing the number of generated conformers to 5. RDKit minimization of the linker fragment after embedding was carried out with the Universal force field~(UFF)~\cite{rappe_uff_1992} with a force convergence criterion of 1e-4 and a energy convergence criterion of 1e-5. Subsequently, the full conformer is minimized using OpenBabel~\cite{openbabel} and the molecular mechanics force field 94~(MMFF94)~\cite{mmff_rdkit} over 500~steps. The steepest descent algorithm is used for minimization. Lastly, each conformer was submitted to Smina minimization~\cite{smina}, which takes the proteins (E3 ligase and POI) into account so as to improve affinity by optimizing the vinardo score and hence also reduce clashing. The best conformer was selected by min-max scaling both the Vinardo score and the RMSD obtained by Smina, and then choosing the best combined score, which was calculated by multiplying the two properties with equal weights. If multiple isomers were enumerated, this approach was applied across all conformers of all isomers.

\section{Deployment of DiffLinker}
\label{sec:SI_difflinker}
5,000 samples were generated using DiffLinker for every investigated system. Since the method does not predict edges, OpenBabel is used to infer bonds as implemented by the method. The original connectivity is kept in the input fragments. In order to circumvent memory issues faced when performing inference on certain systems (5T35, 7JTO, 7JTP, 7Q2J), the input fragments were truncated to smaller substructures containing their respective attachment atoms and atoms or cycles up to 4~hops away from the attachment atoms.

\section{Additional results}
\label{sec:SI_add_results}

\subsection{In-depth evaluation}
This section includes all calculated metrics (\textit{vide infra}) both averaged over all examined structures as well as individually.
\label{sec:SI_detail_analysis}
\begin{table}[H]
	\centering
	\small
	\caption{Chamfer distances between the surface-aligned generated extended linkers and the respective crystal structure pose. To find the best pose, 50~conformers were generated for each linker using RDKit. These results demonstrate the ability of the model to optimize for shape alignment during RL, which was only applied to the new method but not to the baseline Link-INVENT version.}
    \vspace{0.1in}
	\begin{tabularx}{0.79\textwidth}{llllll}
    \toprule
        &&\multicolumn{4}{c}{\textbf{Chamfer distance}}\\\cmidrule{3-6}
		&\textbf{Method}&\textbf{avg $\downarrow$}&  \bm{$<3.5$} \textbf{[\%] $\uparrow$}&\bm{$<2.0$} \textbf{[\%] $\uparrow$}& \bm{$<1.0$} \textbf{[\%] $\uparrow$}\\\midrule
        \multirow{2}{*}{\textbf{all}}&Link-INVENT&4.44&35.83&8.41&0.18\\
		&ShapeLinker&\bm{$2.19$}&\bm{$88.81$}&\bm{$53.93$}&\bm{$2.9$}\\\midrule
        \multirow{2}{*}{5T35}&Link-INVENT&2.16&97.67&41.14&1.25\\
		&ShapeLinker&1.43&99.53&90.33&13.92\\\midrule
        \multirow{2}{*}{7ZNT}&Link-INVENT&5.29&23.19&3.16&0.02\\
		&ShapeLinker&1.47&99.57&87.76&12.18\\\midrule
        \multirow{2}{*}{6HAY}&Link-INVENT&5.74&15.71&0.92&0.00\\
		&ShapeLinker&2.16&98.16&40.84&0.00\\\midrule
        \multirow{2}{*}{6HAX}&Link-INVENT&3.91&39.67&5.59&0.00\\
		&ShapeLinker&1.79&99.73&76.44&0.08\\\midrule
        \multirow{2}{*}{7S4E}&Link-INVENT&4.44&30.38&2.77&0.02\\
		&ShapeLinker&1.58&99.66&90.14&1.14\\\midrule
        \multirow{2}{*}{6BN7}&Link-INVENT&5.08&2.93&0.00&0.00\\
		&ShapeLinker&5.12&1.78&0.00&0.00\\\midrule
        \multirow{2}{*}{6BOY}&Link-INVENT&6.16&9.46&0.34&0.00\\
		&ShapeLinker&2.51&97.03&10.11&0.00\\\midrule
        \multirow{2}{*}{7JTP}&Link-INVENT&5.45&7.13&0.07&0.00\\
		&ShapeLinker&2.10&94.30&46.36&0.13\\\midrule
        \multirow{2}{*}{7Q2J}&Link-INVENT&2.99&69.02&23.71&0.48\\
		&ShapeLinker&1.79&99.11&72.20&1.66\\\midrule
        \multirow{2}{*}{7JTO}&Link-INVENT&3.56&57.12&4.23&0.00\\
		&ShapeLinker&2.24&99.24&24.69&0.00\\\bottomrule
	\end{tabularx}
	\label{tab:SI_RL_cmf_ours_all}
\end{table}

An array of various evaluation metrics are reported. First, measures assessing the generative properties of the methods are calculated according to GuacaMol.\cite{brown_guacamol_2019} These include validity, uniqueness and novelty (with PROTAC-DB as reference), where the latter two do not take stereochemistry into consideration. The highest Tanimoto score to the query PROTAC is listed. \par 
Several metrics evaluating the 3D geometry are reported: the average Chamfer distance~(CD) to the reference crystal structure linker, average root-mean-square deviation~(RMSD) for the anchor and warhead fragments for all constrained embedded conformers (it is zero for the DiffLinker output by design) and the similarity score~$\text{SC}_{\text{RDKit}}$~\cite{scrdkit} to the crystal structure PROTAC assessing topological and chemical similarity. The average CD is of importance as it demonstrates the ability to design linkers of a given shape while the RMSD is hugely impacted by the choice of method for the constrained embedding and thus less insightful. Additionally, a custom shape novelty~(SN) score is introduced, for which the Chamfer distance to the crystal structure linker (inverse min-max scaled) is multiplied by the Tanimoto diversity (1-similarity) score. The average SN captures our main goal of generating topologically similar, but chemically diverse linkers. The torsion energy~($E_{tor}$) determined with OpenBabel~\cite{openbabel} for the whole molecule and the number of clashes with the protein are reported. In accordance with DiffLinker, the ligand clashes with the protein if the distance between a given pair of heavy atoms is bigger than their combined Van der Waals radius. \par 
In addition, properties of particular relevance to the PROTAC drug modality are reported. These include average number of rings, average number of rotational bonds and fraction of branched linkers. The latter two are properties directly optimized for with ShapeLinker and Link-INVENT and these metrics thus further reflect the optimization capability. To assess chemical plausibility of the linker fragment in the context of drug discovery, the average quantitative estimate of drug-likeness~(QED)~\cite{qed_bickerton}, the average SA score~\cite{SA_score} and the fraction passing the 2D filters described in \citet{difflinker} are computed. These include the pan assay interference compounds~(PAINS) filter~\cite{pains} and a ring aromaticity~(RA) filter that ensures rings are either fully aliphatic or aromatic.

\begin{table}[h]
\small
	\caption{Performance metrics evaluating the generative properties of the various methods. Novelty references PROTAC-DB~\cite{protacdb} while maximum Tanimoto similarity (max Tanimoto) observed relates to the reference linker found in the crystal structure. The first group of rows corresponds to the metrics assessed across all investigated systems.}
 \begin{center}
	\begin{tabularx}{0.88\textwidth}{llllll}
    \toprule
		&\textbf{Method}&\textbf{Validity [\%]}& \textbf{Uniqueness [\%]}&\textbf{Novelty [\%]}&\textbf{max Tanimoto $\uparrow$}\\\midrule
        \multirow{3}{*}{\rotatebox[origin=c]{90}{\textbf{all}}}&\cellcolor{gray_others}Link-INVENT&\cellcolor{gray_others}91.65&\cellcolor{gray_others}\bm{$97.39$}&\cellcolor{gray_others}99.94&\cellcolor{gray_others}\bm{$1.00$}\\
        &\cellcolor{gray_others}DiffLinker&\cellcolor{gray_others}70.81&\cellcolor{gray_others}37.85&\cellcolor{gray_others}99.94&\cellcolor{gray_others}\bm{$1.00$}\\\cmidrule{2-6}
		&\cellcolor{gray_shape}ShapeLinker&\cellcolor{gray_shape}\bm{$93.10$}&\cellcolor{gray_shape}95.47&\cellcolor{gray_shape}99.94&\cellcolor{gray_shape}0.91\\\midrule
        \multirow{3}{*}{\rotatebox[origin=c]{90}{5T35}}&Link-INVENT&\bm{$94.46$}&\bm{$95.74$}&100.00&0.19\\
        &DiffLinker&47.40&79.24&100.00&\bm{$1.00$}\\\cmidrule{2-6}
		&ShapeLinker&92.68&92.97&100.00&0.38\\\midrule
        \multirow{3}{*}{\rotatebox[origin=c]{90}{7ZNT}}&Link-INVENT&92.54&\bm{$98.14$}&100.00&0.14\\
        &DiffLinker&76.84&7.11&100.00&\bm{$1.00$}\\\cmidrule{2-6}
		&ShapeLinker&\bm{$93.12$}&88.38&100.00&0.25\\\midrule
        \multirow{3}{*}{\rotatebox[origin=c]{90}{6HAY}}&Link-INVENT&84.52&\bm{$96.99$}&\bm{$100.00$}&0.15\\
        &DiffLinker&86.38&28.13&99.90&\bm{$1.00$}\\\cmidrule{2-6}
		&ShapeLinker&\bm{$94.62$}&95.46&99.98&0.33\\\midrule
        \multirow{3}{*}{\rotatebox[origin=c]{90}{6HAX}}&Link-INVENT&95.14&94.64&99.93&1.00\\
        &DiffLinker&86.46&70.62&99.96&1.00\\\cmidrule{2-6}
		&ShapeLinker&\bm{$95.34$}&\bm{$95.91$}&\bm{$100.00$}&0.42\\\midrule
        \multirow{3}{*}{\rotatebox[origin=c]{90}{7S4E}}&Link-INVENT&\bm{$95.86$}&\bm{$98.5$}&100.00&0.67\\
        &DiffLinker&77.96&64.83&99.91&\bm{$1.00$}\\\cmidrule{2-6}
		&ShapeLinker&93.14&98.37&100.00&0.91\\\midrule
        \multirow{3}{*}{\rotatebox[origin=c]{90}{7JTP}}&Link-INVENT&89.52&\bm{$97.97$}&100.00&0.08\\
        &DiffLinker
        &\bm{$98.02$}&3.92&99.44&\bm{$1.00$}\\\cmidrule{2-6}
		&ShapeLinker&91.98&95.93&100.00&0.56\\\midrule
        \multirow{3}{*}{\rotatebox[origin=c]{90}{7Q2J}}&Link-INVENT&91.28&97.83&100.00&0.41\\
        &DiffLinker
        &93.32&33.82&99.93&\bm{$1.00$}\\\cmidrule{2-6}
		&ShapeLinker&\bm{$94.04$}&\bm{$98.6$}&100.00&0.51\\\midrule
        \multirow{3}{*}{\rotatebox[origin=c]{90}{7JTO}}&Link-INVENT&89.88&99.53&100.00&0.33\\
        &DiffLinker
        &0.06&\bm{$100.00$}&100.00&\bm{$0.63$}\\\cmidrule{2-6}
		&ShapeLinker&\bm{$89.90$}&98.64&100.00&0.5\\\bottomrule
	\end{tabularx}
 \end{center}
	\label{tab:SI_div_metrics_full}
\end{table}

\begin{table}[h!]
	\caption{Performance metrics evaluating the ability to generate linkers that lead to molecules with a close geometry to the reference (Chamfer distance (CD), RMSD and $\text{SC}_{\text{RDKit}}$) as well as a good geometry in relation to the protein (number of clashes (\# Cl)) and energetics (torsion energy in~$[\frac{\text{kcal}}{\text{mol}}]$). The shape novelty (SN) score captures the ability to generate linkers with similar shape but new chemistry. \textit{Fail} reports the fraction that failed constrained embedding resulting $n$ unique samples for which the rest of the metrics were calculated. $\text{DiffLinker}_{\text{CE}}$ refers to conformers obtained by constrained embedding (deduplicated based on SMILES) while $\text{DiffLinker}_{\text{ori}}$ refers to the generated poses with unique conformations but replicate SMILES. The first group of rows corresponds to the metrics assessed across all investigated systems. (anc = anchor, wrh = warhead)}
 \begin{center}
 \begin{small}
    \begin{tabularx}{\textwidth}{lllllllllll}
    \toprule
        &&&&&&\multicolumn{2}{c}{\textbf{RMSD} $\downarrow$}&&&\\\cmidrule{7-8}
    		&\textbf{Method}&\textbf{Fail [\%]} $\downarrow$&\bm{$n$}&\textbf{SN} $\uparrow$&\textbf{CD} $\downarrow$&\textbf{anc}&\textbf{wrh}&\textbf{SC} $\uparrow$&\textbf{\# Cl} $\downarrow$&\bm{$E_{tor}$} $\downarrow$\\\midrule
        \multirow{4}{*}{\rotatebox[origin=c]{90}{\textbf{all}}}&\cellcolor[gray]{0.9}Link-INVENT&\cellcolor{gray_others}27.88&\cellcolor{gray_others}20,967&\cellcolor{gray_others}0.82&\cellcolor{gray_others}5.02&\cellcolor{gray_others}0.56&\cellcolor{gray_others}0.68&\cellcolor{gray_others}0.71&\cellcolor[gray]{0.9}14&\cellcolor{gray_others}69.19\\
        &\cellcolor{gray_others}$\text{DiffLinker}_{\text{CE}}$&\cellcolor{gray_others}3.63&\cellcolor{gray_others}7,936&\cellcolor{gray_others}0.87&\cellcolor{gray_others}1.96&\cellcolor{gray_others}\bm{$0.37$}&\cellcolor{gray_others}\bm{$0.53$}&\cellcolor{gray_others}0.82&\cellcolor{gray_others}\bm{$11$}&\cellcolor{gray_others}\bm{$58.24$}\\
        &\cellcolor{gray_others}$\text{DiffLinker}_{\text{ori}}$&\cellcolor{gray_others}0.00&\cellcolor{gray_others}25,151&\cellcolor{gray_others}0.67&\cellcolor{gray_others}\bm{$1.44$}&\cellcolor{gray_others}-&\cellcolor{gray_others}-&\cellcolor{gray_others}\bm{$0.94$}&\cellcolor{gray_others}13&\cellcolor{gray_others}60.34\\\cmidrule{2-11}
        &\cellcolor{gray_shape}ShapeLinker&\cellcolor{gray_shape}21.45&\cellcolor{gray_shape}14,769&\cellcolor{gray_shape}\bm{$0.9$}&\cellcolor{gray_shape}2.64&\cellcolor{gray_shape}0.47&\cellcolor{gray_shape}0.65&\cellcolor{gray_shape}0.77&\cellcolor{gray_shape}12&\cellcolor{gray_shape}65.62\\\midrule

        \multirow{4}{*}{\rotatebox[origin=c]{90}{5T35}}&Link-INVENT&13.79&1,550&0.89&3.78&1.29&0.98&0.63&14&76.74\\
        &$\text{DiffLinker}_{\text{CE}}$&1.43&1,585&0.86&1.69&\bm{$0.46$}&\bm{$0.51$}&0.80&\bm{$10$}&\bm{$59.93$}\\
        &$\text{DiffLinker}_{\text{ori}}$&0.00&2,095&0.86&\bm{$1.57$}&-&-&\bm{$0.94$}&11&61.31\\\cmidrule{2-11}
		&ShapeLinker&4.80&3,448&\bm{$0.90$}&3.18&0.69&0.93&0.71&11&69.95\\\midrule
  
        \multirow{4}{*}{\rotatebox[origin=c]{90}{7ZNT}}&Link-INVENT&54.15&1,944&0.81&7.40&0.47&0.83&0.69&10&73.45\\
        &$\text{DiffLinker}_{\text{CE}}$&1.97&199&0.71&4.18&0.47&\bm{$0.49$}&0.81&9&60.72\\
        &$\text{DiffLinker}_{\text{ori}}$&0.00&3,579&0.36&\bm{$1.91$}&-&-&\bm{$0.94$}&10&\bm{$51.52$}\\\cmidrule{2-11}
		&ShapeLinker&17.00&942&\bm{$0.89$}&3.61&\bm{$0.43$}&0.86&0.75&9&75.82\\\midrule
  
        \multirow{4}{*}{\rotatebox[origin=c]{90}{6HAY}}&Link-INVENT&0.81&3,432&0.84&6.11&0.31&0.42&0.77&12&80.97\\
        &$\text{DiffLinker}_{\text{CE}}$&2.28&1,028&0.87&1.53&\bm{$0.30$}&\bm{$0.41$}&0.86&10&56.12\\
        &$\text{DiffLinker}_{\text{ori}}$&0.00&4,131&0.83&1.53&-&-&\bm{$0.95$}&11&\bm{$50.77$}\\\cmidrule{2-11}
		&ShapeLinker&6.56&3,917&\bm{$0.94$}&1.73&0.31&0.43&0.84&11&59.2\\\midrule
  
        \multirow{4}{*}{\rotatebox[origin=c]{90}{6HAX}}&Link-INVENT&0.54&4,051&0.77&5.05&0.34&0.62&0.74&10&61.91\\
        &$\text{DiffLinker}_{\text{CE}}$&2.78&1,927&0.89&2.74&0.33&0.51&0.81&9&58.03\\
        &$\text{DiffLinker}_{\text{ori}}$&0.00&3,116&0.87&\bm{$2.31$}&-&-&\bm{$0.91$}&\bm{$6$}&\bm{$55.79$}\\\cmidrule{2-11}
		&ShapeLinker&7.17&1,889&0.89&3&0.33&0.51&0.81&9&56.44\\\midrule
  
        \multirow{4}{*}{\rotatebox[origin=c]{90}{7S4E}}&Link-INVENT&0.24&3,792&0.74&6.21&0.36&0.61&0.73&12&56.59\\
        &$\text{DiffLinker}_{\text{CE}}$&2.23&1,884&\bm{$0.91$}&1.85&0.35&0.69&0.8&11&53.31\\
        &$\text{DiffLinker}_{\text{ori}}$&0.00&3,197&0.88&\bm{$1.48$}&-&-&\bm{$0.94$}&15&\bm{$48.22$}\\\cmidrule{2-11}
		&ShapeLinker&38.45&586&0.87&2.17&0.35&\bm{$0.59$}&0.81&11&56.88\\\midrule
  
        \multirow{4}{*}{\rotatebox[origin=c]{90}{7JTP}}&Link-INVENT&98.8&49&0.85&5.77&1&0.67&0.68&37&89.17\\
        &$\text{DiffLinker}_{\text{CE}}$&59.12&65&\bm{$0.91$}&0.86&\bm{$0.45$}&\bm{$0.48$}&0.83&19&90.75\\
        &$\text{DiffLinker}_{\text{ori}}$&0.00&4,741&0.4&\bm{$0.59$}&-&-&\bm{$0.98$}&\bm{$16$}&\bm{$85.90$}\\\cmidrule{2-11}
		&ShapeLinker&96.44&34&0.89&2.35&0.76&0.5&0.76&23&100.89\\\midrule
  
        \multirow{4}{*}{\rotatebox[origin=c]{90}{7Q2J}}&Link-INVENT&7.44&3,558&0.88&3.87&0.91&0.92&0.65&23&68.52\\
        &$\text{DiffLinker}_{\text{CE}}$&4.30&1,245&0.79&1.32&\bm{$0.39$}&\bm{$0.42$}&0.84&\bm{$13$}&63.54\\
        &$\text{DiffLinker}_{\text{ori}}$&0.00&4,289&0.70&\bm{$1.17$}&-&-&\bm{$0.94$}&22&\bm{$60.56$}\\\cmidrule{2-11}
		&ShapeLinker&20.60&1,950&0.88&1.57&0.55&0.52&0.77&17&68.84\\\midrule
  
        \multirow{4}{*}{\rotatebox[origin=c]{90}{7JTO}}&Link-INVENT&31.44&2,591&\bm{$0.89$}&2.34&0.63&0.62&0.70&17&76.25\\
        &$\text{DiffLinker}_{\text{CE}}$&0.00&3&0.66&2.31&\bm{$0.41$}&\bm{$0.5$}&0.77&\bm{$12$}&66.51\\
        &$\text{DiffLinker}_{\text{ori}}$&0.00&3&0.66&\bm{$2.22$}&-&-&\bm{$0.88$}&20&\bm{$66.39$}\\\cmidrule{2-11}
        &ShapeLinker&42.03&2,003&0.84&3.84&0.52&0.74&0.74&14&73.33\\\bottomrule
	\end{tabularx}
\end{small}
\end{center}
	\label{tab:SI_geom_metrics_full}
\end{table}

\begin{table}[h]
 \begin{center}
 \begin{small}
	\caption{ Performance metrics assessing the drug-likeness (QED) and synthesizability (SA) of the generated molecules and the chemical suitability specifically to the class of PROTAC drugs (number of rings, number of rotational bonds (ROT) and fraction of branched linkers). All metrics reference the linker fragment only, except for the PAINS filter within the 2D Filters, which is used to identify problematic new connections. The first group of rows corresponds to the metrics assessed across all investigated systems.}
	\vspace{0.1in}
    \begin{tabularx}{\textwidth}{lllllllll}
    \toprule
		&\textbf{Method}&\bm{$n$}&\textbf{QED $\uparrow$}&\textbf{SA $\downarrow$}&\textbf{Filters [\%] $\uparrow$}&\textbf{\#Rings $\uparrow$}&\textbf{\#ROT $\downarrow$}&\textbf{Branch [\%] $\downarrow$}\\\midrule
		\multirow{3}{*}{\rotatebox[origin=c]{90}{\textbf{all}}}&\cellcolor{gray_others}Link-INVENT&\cellcolor{gray_others}36,660&\cellcolor{gray_others}\bm{$0.66$}&\cellcolor{gray_others}2.98&\cellcolor{gray_others}92.83&\cellcolor{gray_others}\bm{$1.98$}&\cellcolor{gray_others}3.27&\cellcolor{gray_others}12.06\\
        &\cellcolor{gray_others}DiffLinker&\cellcolor{gray_others}28,322&\cellcolor{gray_others}0.5&\cellcolor{gray_others}\bm{$2.55$}&\cellcolor{gray_others}\bm{$94.32$}&\cellcolor{gray_others}0.32&\cellcolor{gray_others}2.60&\cellcolor{gray_others}9.66\\\cmidrule{2-9}
		&\cellcolor{gray_shape}ShapeLinker&\cellcolor{gray_shape}37,241&\cellcolor{gray_shape}0.51&\cellcolor{gray_shape}3.74&\cellcolor{gray_shape}76.51&\cellcolor{gray_shape}0.91&\cellcolor{gray_shape}\bm{$1.67$}&\cellcolor{gray_shape}\bm{$8.64$}\\\midrule
        \multirow{3}{*}{\rotatebox[origin=c]{90}{5T35}}&Link-INVENT&4,723&0.52&4.12&\bm{$96.99$}&\bm{$1.64$}&\bm{$1.68$}&19.65\\
        &DiffLinker&2,370&0.53&\bm{$2.69$}&94.56&0.31&4.47&15.74\\\cmidrule{2-9}
		&ShapeLinker&4,634&\bm{$0.57$}&3.13&93.7&1.03&2.56&\bm{$1.77$}\\\midrule
        \multirow{3}{*}{\rotatebox[origin=c]{90}{7ZNT}}&Link-INVENT&4,627&\bm{$0.71$}&2.46&\bm{$95.35$}&\bm{$1.79$}&4.70&2.46\\
        &DiffLinker&3,842&0.47&\bm{$1.68$}&93.49&0.03&2.76&3.62\\\cmidrule{2-9}
		&ShapeLinker&4,656&0.44&4.15&55.84&0.99&\bm{$1.07$}&\bm{$1.91$}\\\midrule
        \multirow{3}{*}{\rotatebox[origin=c]{90}{6HAY}}&Link-INVENT&4,226&\bm{$0.73$}&3.06&89.99&\bm{$3.03$}&3.53&7.76\\
        &DiffLinker&4,319&0.52&\bm{$2.31$}&98.43&0.15&3.79&\bm{$7.73$}\\\cmidrule{2-9}
		&ShapeLinker&4,731&0.62&2.92&\bm{$98.69$}&1.04&\bm{$2.14$}&10.80\\\midrule
        \multirow{3}{*}{\rotatebox[origin=c]{90}{6HAX}}&Link-INVENT&4,757&\bm{$0.73$}&\bm{$2.22$}&\bm{$93.17$}&\bm{$2.08$}&2.68&8.62\\
        &DiffLinker&4,323&0.52&3.06&84.50&0.94&1.74&16.59\\\cmidrule{2-9}
		&ShapeLinker&4,767&0.52&3.85&77.43&1.08&\bm{$0.75$}&\bm{$4.64$}\\\midrule 
        \multirow{3}{*}{\rotatebox[origin=c]{90}{7S4E}}&Link-INVENT&4,793&\bm{$0.71$}&3.00&87.54&\bm{$2.08$}&3.99&6.36\\
        &DiffLinker&3,898&0.56&\bm{$2.77$}&\bm{$89.66$}&0.65&3.54&11.54\\\cmidrule{2-9}
		&ShapeLinker&4,657&0.49&4.17&39.96&0.88&\bm{$1.12$}&\bm{$7.00$}\\\midrule
        \multirow{3}{*}{\rotatebox[origin=c]{90}{7JTP}}&Link-INVENT&4,476&\bm{$0.73$}&\bm{$2.77$}&96.49&\bm{$1.98$}&2.72&28.42\\
        &DiffLinker&4,901&0.41&2.79&\bm{$99.9$}&0.02&\bm{$0.83$}&\bm{$6.16$}\\\cmidrule{2-9}
		&ShapeLinker&4,599&0.40&4.54&68.82&0.44&1.59&9.98\\\midrule
        \multirow{3}{*}{\rotatebox[origin=c]{90}{7Q2J}}&Link-INVENT&4,564&\bm{$0.64$}&3.27&93.16&\bm{$1.48$}&2.73&11.77\\
        &DiffLinker&4,666&0.5&\bm{$2.54$}&\bm{$98.18$}&0.19&2.32&\bm{$9.00$}\\\cmidrule{2-9}
		&ShapeLinker&4,702&0.50&3.84&88.88&0.61&\bm{$1.91$}&27.86\\\midrule
        \multirow{3}{*}{\rotatebox[origin=c]{90}{7JTO}}&Link-INVENT&4,494&0.53&2.93&89.79&\bm{$1.84$}&4.18&11.77\\
        &DiffLinker&3&0.46&\bm{$2.50$}&\bm{$100.00$}&0.33&8.00&33.33\\\cmidrule{2-9}
		&ShapeLinker&4,495&\bm{$0.57$}&3.3&88.68&1.18&\bm{$2.25$}&\bm{$4.85$}\\\bottomrule
	\end{tabularx}
\end{small}
\end{center}
	\label{tab:SI_chem_metrics_full}
\end{table}

To assess the differences in the poses directly obtained from the method and the constrained embedded poses, the Chamfer distances between each pair was calculated and a summary is listed in Table~\ref{tab:SI_embedded2method}. Overall, the poses generated with DiffLinker and the shape-aligned poses from ShapeLinker are equally comparable to the constrained embedded poses, while Link-INVENT results in substantially larger Chamfer distances. 

\begin{table}[h]
	\centering
	\small
	\caption{Aligned chamfer distances between the linker conformation resulting from either method and the respective poses obtained by constrained embedding. The structure used for chamfer distance calculation refers to the surface aligned linker for our work, while for DiffLinker, the predicted pose is used.}
    \vspace{0.1in}
	\begin{tabularx}{0.79\textwidth}{llllll}
    \toprule
        &&\multicolumn{4}{c}{\textbf{Chamfer distance}}\\\cmidrule{3-6}
		&\textbf{Method}&\textbf{avg $\downarrow$}&  \bm{$<3.5$} \textbf{[\%] $\uparrow$}&\bm{$<2.0$} \textbf{[\%] $\uparrow$}& \bm{$<1.0$} \textbf{[\%] $\uparrow$}\\\midrule
        \multirow{3}{*}{\textbf{all}}&\cellcolor{gray_others}Link-INVENT&\cellcolor{gray_others}2.25&\cellcolor{gray_others}85.48&\cellcolor{gray_others}51.57&\cellcolor{gray_others}11.6\\
        &\cellcolor{gray_others}DiffLinker&\cellcolor{gray_others}\bm{$1.23$}&\cellcolor{gray_others}\bm{$97.35$}&\cellcolor{gray_others}\bm{$92.20$}&\cellcolor{gray_others}\bm{$50.56$}\\\cmidrule{2-6}
		&\cellcolor{gray_shape}ShapeLinker&\cellcolor{gray_shape}1.30&\cellcolor{gray_shape}97.08&\cellcolor{gray_shape}88.16&\cellcolor{gray_shape}42.56\\\midrule
        \multirow{3}{*}{5T35}&Link-INVENT&\textbf{$1.78$}&\bm{$92.25$}&\bm{$69.72$}&\bm{$22.98$}\\
        &DiffLinker&2.04&87.07&71.8&22.21\\\cmidrule{2-6}
		&ShapeLinker&2.03&87.71&60.64&15.71\\\midrule
        \multirow{3}{*}{7ZNT}&Link-INVENT&2.06&86.99&59.62&19.03\\
        &DiffLinker&\bm{$0.78$}&100.00&\bm{$100.00$}&\bm{$88.94$}\\\cmidrule{2-6}
		&ShapeLinker&0.87&100.00&99.36&73.14\\\midrule
        \multirow{3}{*}{6HAY}&Link-INVENT&2.79&74.30&32.40&7.52\\
        &DiffLinker&\bm{$1.02$}&\bm{$100.00$}&\bm{$98.83$}&\bm{$53.6$}\\\cmidrule{2-6}
		&ShapeLinker&1.10&99.97&96.43&48.86\\\midrule
        \multirow{3}{*}{6HAX}&Link-INVENT&2.09&89.26&56.3&12.94\\
        &DiffLinker&\bm{$0.94$}&99.84&98.34&\bm{$65.75$}\\\cmidrule{2-6}
		&ShapeLinker&0.95&\bm{$100.00$}&\bm{$98.78$}&65.17\\\midrule
        \multirow{3}{*}{7S4E}&Link-INVENT&2.56&81.22&39.71&5.15\\
        &DiffLinker&1.23&100.00&93.42&37.26\\\cmidrule{2-6}
		&ShapeLinker&\bm{$1.05$}&100.00&\bm{$97.95$}&\bm{$53.24$}\\\midrule
        \multirow{3}{*}{7JTP}&Link-INVENT&1.54&100.00&81.25&22.92\\
        &DiffLinker&\bm{$0.42$}&100.00&100.00&\bm{$100.00$}\\\cmidrule{2-6}
		&ShapeLinker&0.99&100.00&100.00&61.76\\\midrule
        \multirow{3}{*}{7Q2J}&Link-INVENT&1.85&93.07&67.32&14.08\\
        &DiffLinker&0.92&100.00&\bm{$100.00$}&\bm{$72.29$}\\\cmidrule{2-6}
		&ShapeLinker&\bm{$1.14$}&100.00&97.37&38.85\\\midrule
        \multirow{3}{*}{7JTO}&Link-INVENT&2.35&84.78&47.89&8.30\\
        &DiffLinker&3.85&33.33&0.00&0.00\\\cmidrule{2-6}
		&ShapeLinker&\bm{$1.22$}&\bm{$99.65$}&\bm{$92.01$}&\bm{$40.79$}\\\bottomrule
	\end{tabularx}
	\label{tab:SI_embedded2method}
\end{table}

\clearpage
\newpage
\subsection{Visualization of selected generated examples}

\begin{figure}[h]
    \centering
    \includegraphics[width=\textwidth]{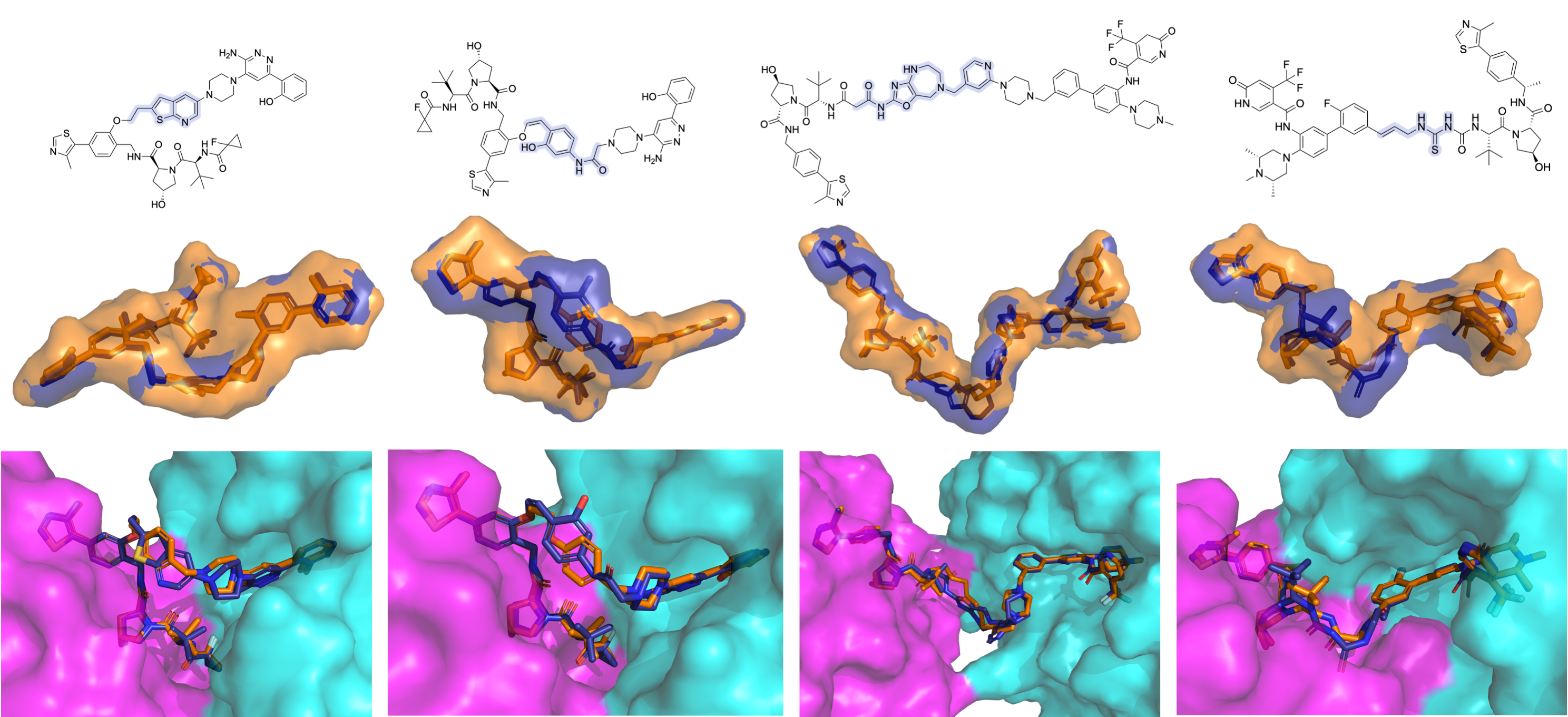}
    \caption{Selected examples of samples generated by ShapeLinker (dark blue) compared to their respective crystal structure PROTAC (orange). The upper images show the 2D structures with highlighted linker fragment the middle row shows the aligned surfaces of the reference (orange) and generated PROTAC (blue) and the lower images show the 3D structures binding the E3 ligase (pink) and the POI (light blue). Examples from left to right: 6HAX, 7S4E, 7JTO, 7JTP.}
    \label{fig:SI_example_structures}
\end{figure}

\begin{figure}[h]
\centering
\subfigure{\includegraphics[width=\textwidth]{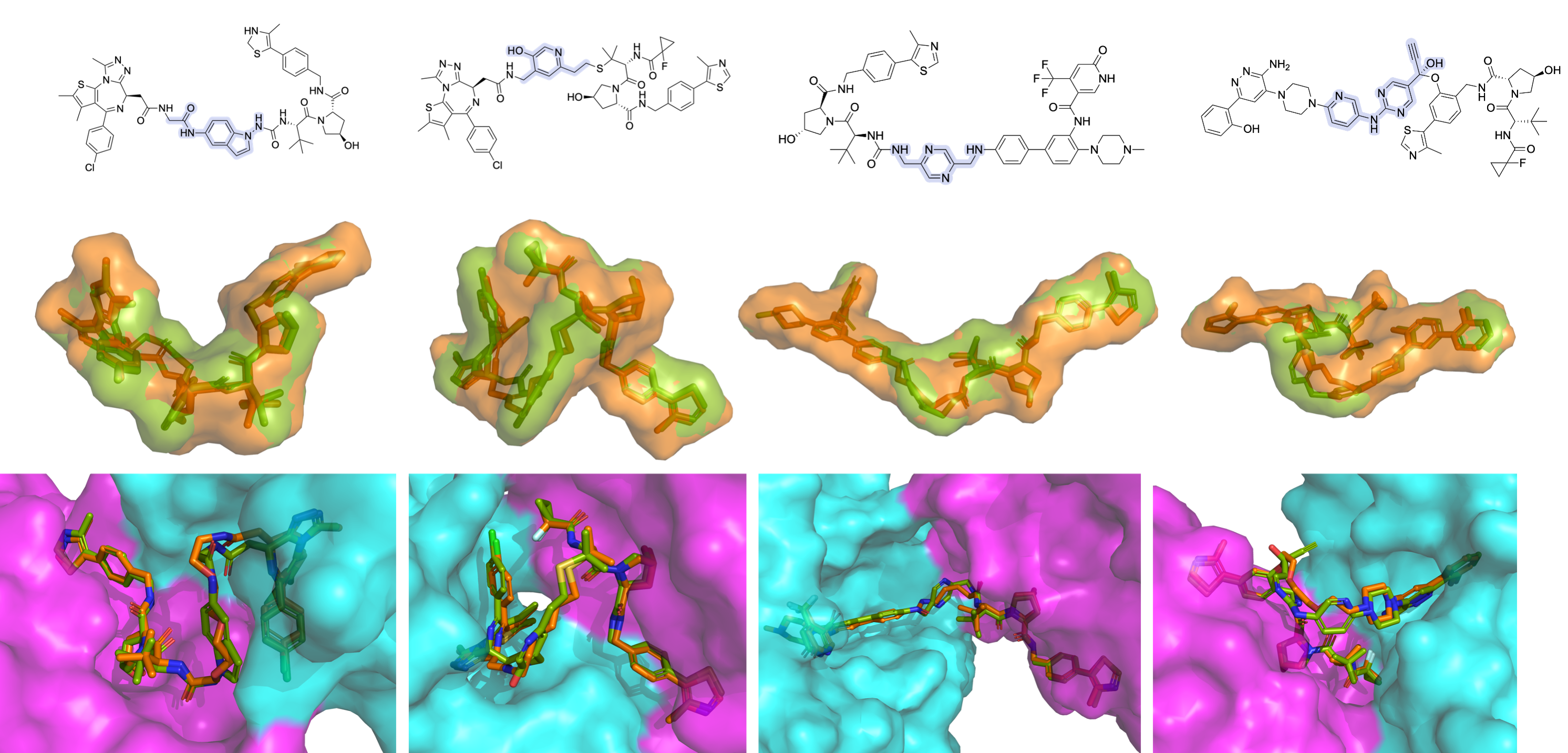}}
\vspace{0.1cm}
\hrule
\subfigure{\includegraphics[width=\textwidth]{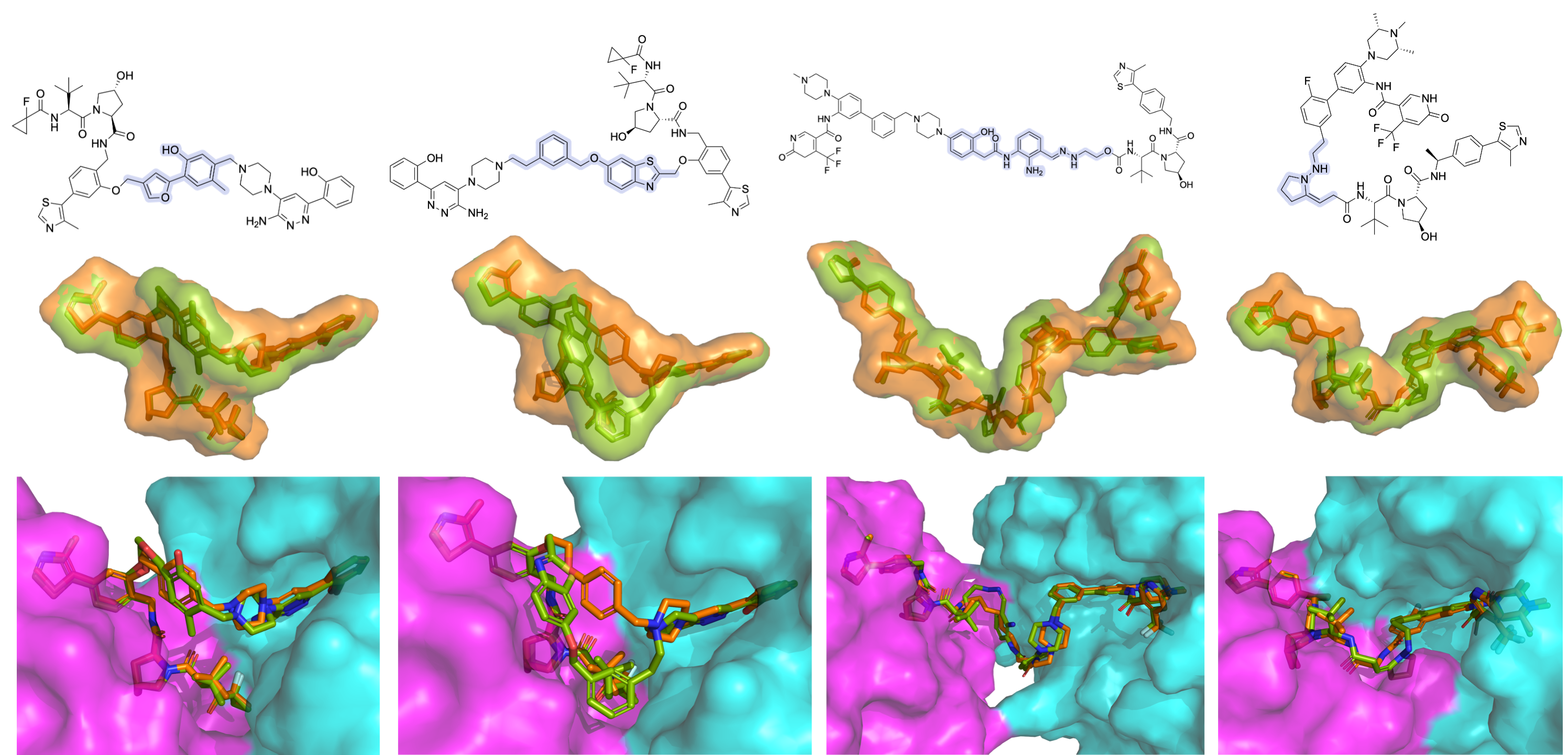}}
\caption{Selected examples of samples generated by Link-INVENT (green) compared to their respective crystal structure PROTAC (orange). The upper images show the 2D structures with highlighted linker fragment the middle row shows the aligned surfaces of the reference (orange) and generated PROTAC (blue) and the lower images show the 3D structures binding the E3 ligase (pink) and the POI (light blue). Examples from left to right: \textit{upper row}: 5T35, 7S4E, 7JTO, 7JTP, \textit{lower row}: 6HAX, 7S4E, 7JTO, 7JTP.} \label{fig:SI_base_example_structures}
\end{figure}

\begin{figure}[h]
\centering
\subfigure{\includegraphics[width=\textwidth]{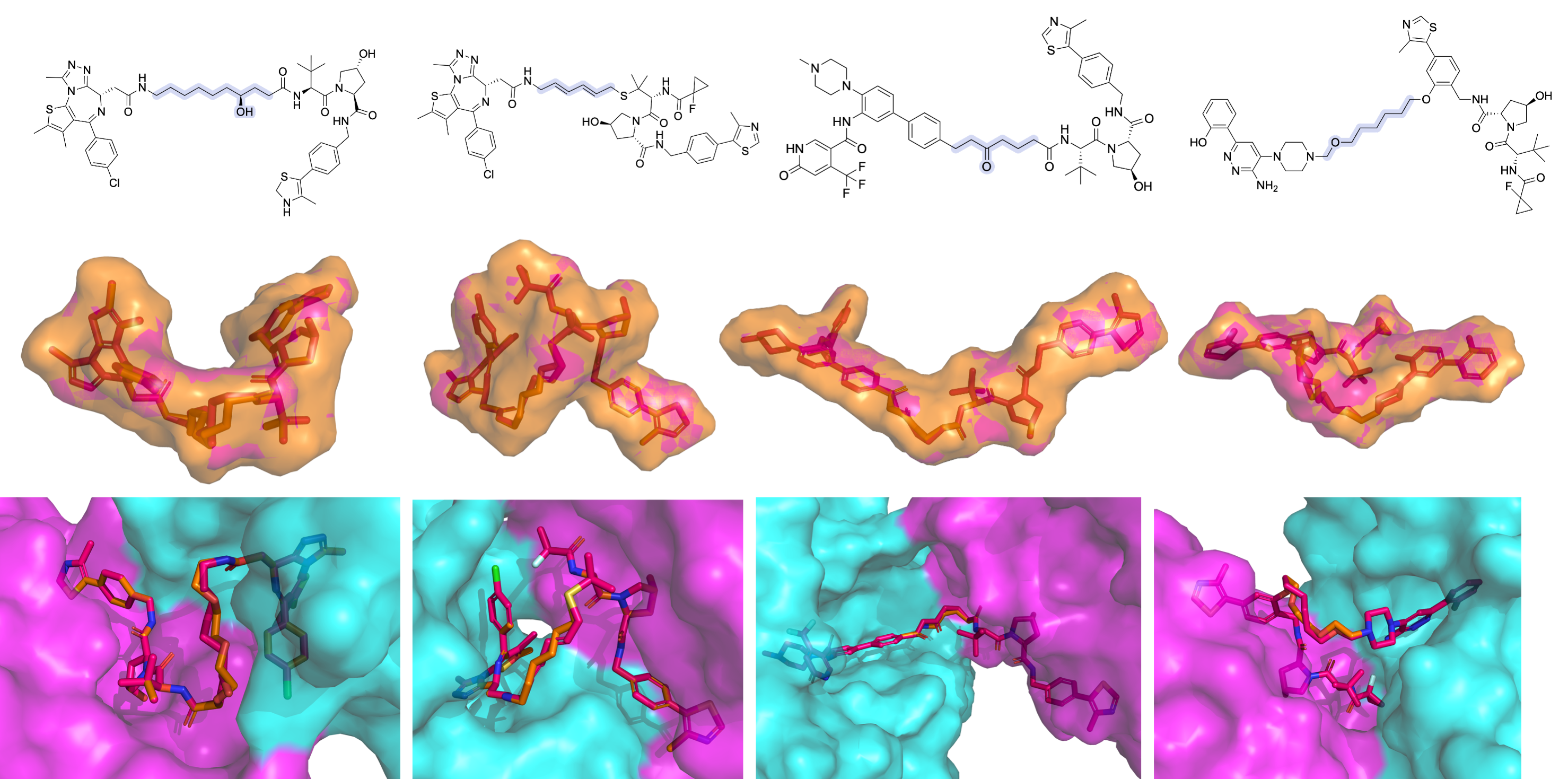}}
\vspace{0.1cm}
\hrule
\subfigure{\includegraphics[width=\textwidth]{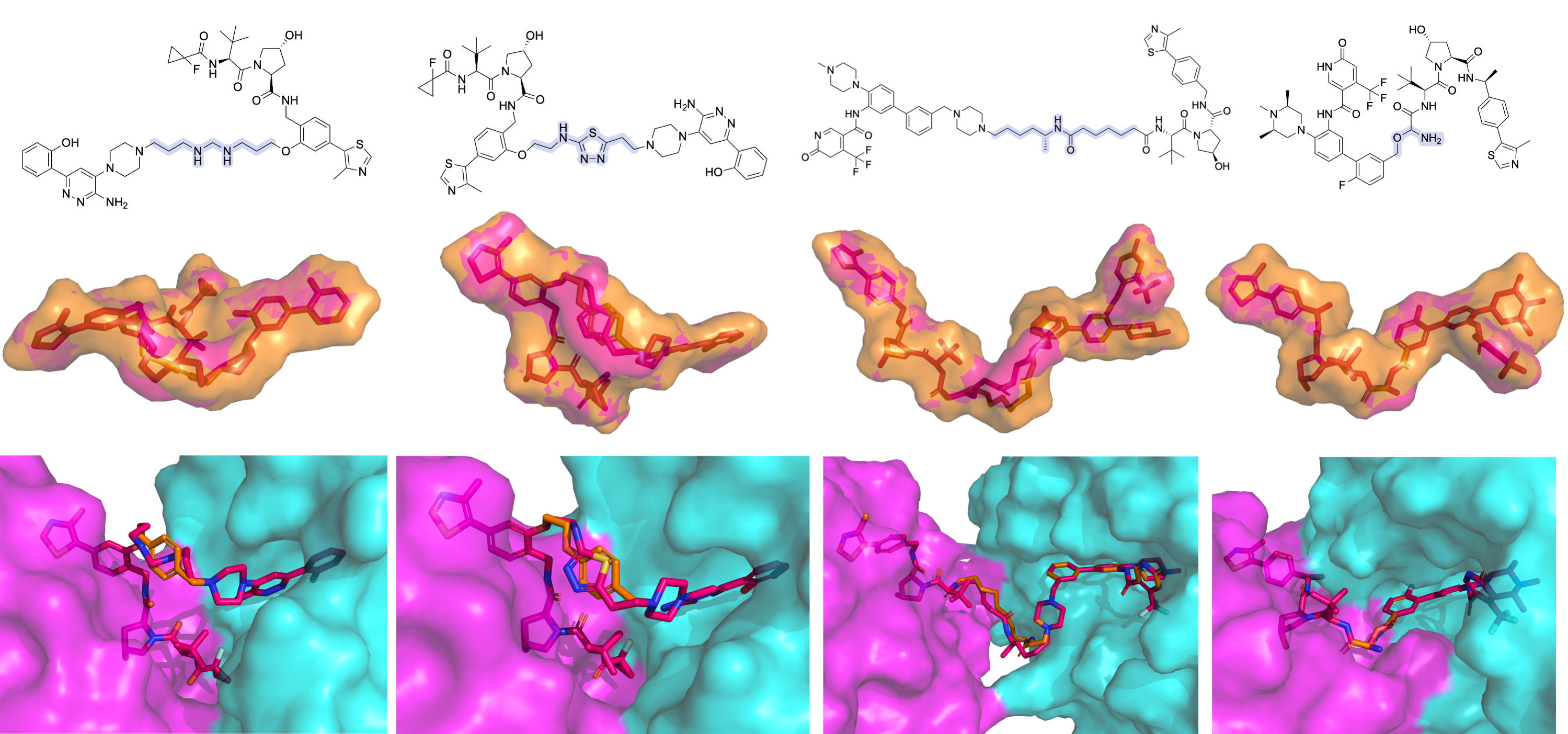}}
\caption{Selected examples of samples generated by DiffLinker (pink) compared to their respective crystal structure PROTAC (orange). The upper images show the 2D structures with highlighted linker fragment the middle row shows the aligned surfaces of the reference (orange) and generated PROTAC (blue) and the lower images show the 3D structures binding the E3 ligase (pink) and the POI (light blue). Examples from left to right: \textit{upper row}: 5T35, 7S4E, 7JTO, 7JTP, \textit{lower row}: 6HAX, 7S4E, 7JTO, 7JTP.} \label{fig:SI_difflinker_example_structures}
\end{figure}

\clearpage
\newpage
\subsection{Results for 6BOY and 6BN7}
\label{sec:SI_6boy_6bn7}
\begin{table}[h]
	\centering
	\small
	\caption{Performance metrics evaluating the generative properties of the various methods. Novelty references PROTAC-DB, while recovery and maximum Tanimoto score (max Tanimoto) observed relates to the reference linker found in the crystal structure.\cite{protacdb}}
    \vspace{0.1in}
	\begin{tabularx}{0.9\textwidth}{ll@{\hskip 0.3in}llll}
    \toprule
		&\textbf{Method}&\textbf{Validity [\%]}& \textbf{Uniqueness [\%]}&\textbf{Novelty [\%]}&\textbf{max Tanimoto $\uparrow$}\\\midrule
        \multirow{3}{*}{\rotatebox[origin=c]{90}{6BN7}}&Link-INVENT&90.86&84.55&100.00&0.34\\
        &DiffLinker&0.06&100.00&100.00&0.28\\\cmidrule{2-6}
		&ShapeLinker&93.36&93.32&100.00&0.43\\\midrule
        \multirow{3}{*}{\rotatebox[origin=c]{90}{6BOY}}&Link-INVENT&83.28&99.93&100.00&0.18\\
        &DiffLinker&0.00&-&-&-\\\cmidrule{2-6}
		&ShapeLinker&94.98&96.88&100.00&0.93\\\midrule
	\end{tabularx}
	\label{tab:SI_div_fails}
\end{table}

\begin{table}[h]
	\centering
	\small
	\caption{Performance metrics assessing the drug-likeness (QED) and synthesizability (SA) of the generated molecules and the chemical suitability specifically to the class of PROTAC drugs (number of rings, number of rotational bonds (ROT) and fraction of branched linkers). All metrics reference the linker fragment only, except for the PAINS filter within the 2D Filters, which is used to identify problematic new connections.}
	\vspace{0.1in}
    \begin{tabularx}{\textwidth}{lllllllll}
    \toprule
		&\textbf{Method}&\bm{$n$}&\textbf{QED $\uparrow$}&\textbf{SA $\downarrow$}&\textbf{Filters [\%] $\uparrow$}&\textbf{\#Rings $\uparrow$}&\textbf{\#ROT $\downarrow$}&\textbf{Branch [\%] $\downarrow$}\\\midrule
        \multirow{3}{*}{\rotatebox[origin=c]{90}{6BN7}}&Link-INVENT&4,543&0.67&2.05&96.65&1.52&2.73&18.51\\
        &DiffLinker&3&0.66&3.1&80.00&0.80&6.20&20.00\\\cmidrule{2-9}
		&ShapeLinker&4,668&0.56&3.17&98.37&1.18&1.82&15.55\\\midrule
        \multirow{3}{*}{\rotatebox[origin=c]{90}{6BOY}}&Link-INVENT&4,164&0.4&2.83&92.12&2.84&7.40&19.6\\
        &DiffLinker&0&-&-&-&-&-&-\\\cmidrule{2-9}
		&ShapeLinker&4,749&0.67&2.75&94.55&1.25&3&5.18\\\midrule
	\end{tabularx}
	\label{tab:SI_chem_fails}
\end{table}

\begin{table}[h]
	\centering
	\small
	\caption{Performance metrics evaluating the ability to generate linkers that lead to molecules with a close geometry to the reference (Chamfer distance (CD), RMSD and $\text{SC}_{\text{RDKit}}$) as well as a good geometry in relation to the protein (number of clashes (\# Cl)) and energetics (torsion energy). The shape novelty (SN) score captures the ability to generate linkers with similar shape but new chemistry. \textit{Fail} reports the fraction that failed constrained embedding resulting $n$ unique samples for which the rest of the metrics were calculated. $\text{DiffLinker}_{\text{CE}}$ refers to conformers obtained by constrained embedding (deduplicated based on SMILES) while $\text{DiffLinker}_{\text{ori}}$ refers to the generated poses with unique conformations but replicate SMILES. Only DiffLinker samples for 6BN7 resulted in productive poses while none of the methods achieved the generation of 3D conformers for 6BOY. (anc = anchor, wrh = warhead)}
	\vspace{0.1in}
    \begin{tabularx}{\textwidth}{lllllllllll}
    \toprule
        &&&&\multicolumn{2}{c}{\textbf{RMSD} $\downarrow$}&&&&&\\\cmidrule{5-6}
    		&\textbf{Method}&\textbf{Fail [\%]} $\downarrow$&\bm{$n$}&\textbf{anc}&\textbf{wrh}&\textbf{CD} $\downarrow$&\textbf{SC} $\uparrow$&\textbf{\# Cl} $\downarrow$&\bm{$E_{tor}$} $[\frac{\text{kcal}}{\text{mol}}]$ $\downarrow$&\textbf{SN} $\uparrow$\\\midrule
        \multirow{4}{*}{\rotatebox[origin=c]{90}{6BN7}}&Link-INVENT&100.00&0&-&-&-&-&-&-&-\\
        &$\text{DiffLinker}_{\text{CE}}$&0.00&3&0.57&1.06&2.4&0.61&11&58.13&0.79\\
        &$\text{DiffLinker}_{\text{ori}}$&0.00&3&-&-&1.99&0.82&24&33.13&0.81\\\cmidrule{2-11}
		&ShapeLinker&100.00&0&-&-&-&-&-&-&-\\\midrule
	\end{tabularx}
	\label{tab:SI_geom_fails}
\end{table}

\end{document}